\documentclass{article} 
\usepackage{iclr2027_conference,times}

\usepackage{hyperref}
\usepackage{url}
\usepackage{booktabs}
\usepackage{multirow}
\usepackage{amssymb}   
\usepackage{amsmath}
\usepackage{pifont}    
\usepackage{array}
\usepackage{xcolor}
\usepackage{graphicx}
\usepackage{cleveref}
\usepackage{makecell}
\usepackage{tabularx}
\usepackage{tikz}

\usepackage{wrapfig}
\usepackage{float}      
\usepackage{caption}    
\usepackage{subcaption} 

\usepackage{listings}
\usepackage{xcolor}
\usepackage{upquote}   

\definecolor{reprnum}{RGB}{0,128,0}      
\definecolor{reprkey}{RGB}{0,0,160}      
\definecolor{reprcom}{RGB}{128,128,128}  

\lstdefinestyle{pytorchrepr}{
  basicstyle=\ttfamily\scriptsize,
  columns=fullflexible,
  keepspaces=true,
  upquote=true,
  breaklines=false,
  showstringspaces=false,
  emph={True,False},
  emphstyle=\bfseries\color{reprkey},
  comment=[l]{\#},
  commentstyle=\color{reprcom},
  literate=%
    {0}{{\color{reprnum}0}}1 {1}{{\color{reprnum}1}}1 {2}{{\color{reprnum}2}}1
    {3}{{\color{reprnum}3}}1 {4}{{\color{reprnum}4}}1 {5}{{\color{reprnum}5}}1
    {6}{{\color{reprnum}6}}1 {7}{{\color{reprnum}7}}1 {8}{{\color{reprnum}8}}1
    {9}{{\color{reprnum}9}}1,
}

\newcommand{\cmark}{\checkmark}
\newcommand{\xmark}{\ding{55}}

\newcommand{\echotab}{\textsc{EchoTab}}

\definecolor{upgreen}{RGB}{26,140,90}
\definecolor{downred}{RGB}{214,45,32}

\title{EchoTab: Learning Measurement-Grounded Representations for Echocardiographic Diagnosis through Tabular Reconstruction}

\author{%
Michelle Espranita Liman\textsuperscript{1},
Ioana Ionac\textsuperscript{2},
Alexander Müller\textsuperscript{2}, \\
\bfseries
Eimo Martens\textsuperscript{2},
Daniel Rueckert\textsuperscript{1,3,4},  
Philip Müller\textsuperscript{1} \\[0.5em]
\parbox{0.97\textwidth}{\small\normalfont
\textsuperscript{1}Chair for AI in Healthcare and Medicine, Technical University of Munich (TUM) and TUM University Hospital, Munich, Germany \\
\textsuperscript{2}Department of Cardiology, TUM University Hospital, Munich, Germany \\
\textsuperscript{3}Department of Computing, Imperial College London, London, UK \\
\textsuperscript{4}Munich Center for Machine Learning (MCML), Munich, Germany \\[0.4em]
Correspondence to: \texttt{michelle.liman@tum.de}
}
}

\iclrfinalcopy 
\begin{document}

\maketitle
\lhead{\textit{Preprint.}}

\begin{abstract}

Diagnostic information in multi-view medical imaging studies, like echocardiography, is distributed across views, each capturing different aspects of cardiac structure and function. In clinical practice, these views are first interpreted through quantitative measurements, which then support specific diagnoses. However, existing echo representation learning methods do not explicitly model measurements as an intermediate representation between views and diagnosis. As a result, individual measurements are not explicitly represented in a form that downstream diagnostic tasks can selectively use, hindering accurate diagnosis.

Inspired by this hierarchical clinical workflow, we introduce \echotab{}, a measurement-grounded representation learning framework that represents multi-view echocardiography studies using measurement-specific tokens as an intermediate representation between echo views and diagnosis. Each measurement token selectively attends to the views relevant to its estimation and is trained through tabular reconstruction to predict the corresponding measurement value. The resulting tokens form a factorized echo representation, which encode clinically meaningful structural and functional properties and can be selectively used for downstream diagnosis. Unlike the commonly used contrastive approaches, \echotab{} does not require negative samples or large batch sizes. Experiments on two datasets demonstrate that \echotab{} learns effective measurement-grounded representations that improve diagnosis classification in multi-view echocardiography and its learned attention aligns with established measurement-view relationships used by cardiologists. Our code is accessible at \url{https://github.com/michelleespranita/EchoTab}.

\end{abstract}

\section{Introduction}
\label{introduction}

Echocardiography (echo) is a common cardiac modality to assess the structure and function of the heart. Echocardiography interpretation follows a hierarchical process in which echo views in an echo study give rise to quantitative measurements that inform clinical diagnosis. Importantly, different measurements are derived from different subsets of views, and different diagnoses are informed by specific subsets of measurements. For example, the left ventricular ejection fraction measurement is derived from apical two- (A2C) and four-chamber (A4C) views and is then used to assess the severity of left ventricular systolic dysfunction.

Despite this hierarchical structure, existing echo foundation models are trained directly \citep{panecho} or must be finetuned \citep{echojepa, echovisionfm} for diagnosis prediction, or leverage retrieval of text reports that contain diagnoses \citep{echoprime}. However, categorical diagnosis labels introduce a discontinuity into the learning target and lose information about how strongly a patient expresses a given phenotype. Instead, measurements provide continuous, dense supervision. Rather than requiring the model to learn a thresholded clinical decision, measurement prediction encourages it to capture the anatomical and functional properties that underlie that decision. Some existing approaches already incorporate measurements but do not explicitly represent them as continuous targets. For example, EchoPrime \citep{echoprime} uses echo reports for contrastive learning, indirectly exposing the model to measurements described in the text. However, these measurements are encoded by the text encoder as discrete text representations rather than as continuous values. A natural alternative is to use measurements directly as the target modality for contrastive learning. However, contrastive learning requires the set of measurements to be encoded into a single tabular embedding, which is then aligned with a single echo embedding. While this makes the resulting echo representations measurement-aware, the individual measurements are not explicitly represented and therefore cannot be factorized within the learned representations.

Motivated by the clinical hierarchy of echocardiographic interpretation, we introduce \echotab{}, a measurement-grounded representation learning framework that models measurements as an intermediate representation between echo views and downstream diagnosis. Rather than collapsing measurements into a single embedding, \echotab{} assigns a dedicated token to each measurement, each selectively attending to the relevant echo views. These measurement-specific tokens thus form a factorized representation of the echo study. A prediction head is trained to reconstruct the corresponding measurements from these tokens, encouraging each token to capture clinically meaningful structural and functional properties. These measurement-specific tokens can then be selectively combined for accurate downstream diagnosis. Unlike contrastive approaches, \echotab{} uses a tabular reconstruction objective, which does not require negative samples and large batch sizes.

Our contributions are:

\begin{itemize}

\item We propose \echotab{}, a measurement-grounded representation learning framework that represents multi-view echocardiography studies using measurement-specific tokens, which serve as an intermediate representation between echo views and diagnosis.

\item We show that \echotab{} learns these measurement-specific tokens through tabular reconstruction. This enables diagnosis-specific selection of relevant measurements, improving diagnosis performance on two datasets over pre-training methods that learn generic study-level representations, while avoiding the need for negative samples and large batch sizes.

\item We systematically study \echotab{}'s architecture and behavior, including the impact of different measurement prediction heads, and show that its learned attention over echo views is consistent with established measurement-view relationships used by cardiologists.

\end{itemize}

\section{EchoTab}
\label{method}

\begin{figure}[t]
    \centering
        \centering
        \includegraphics[width=\linewidth]{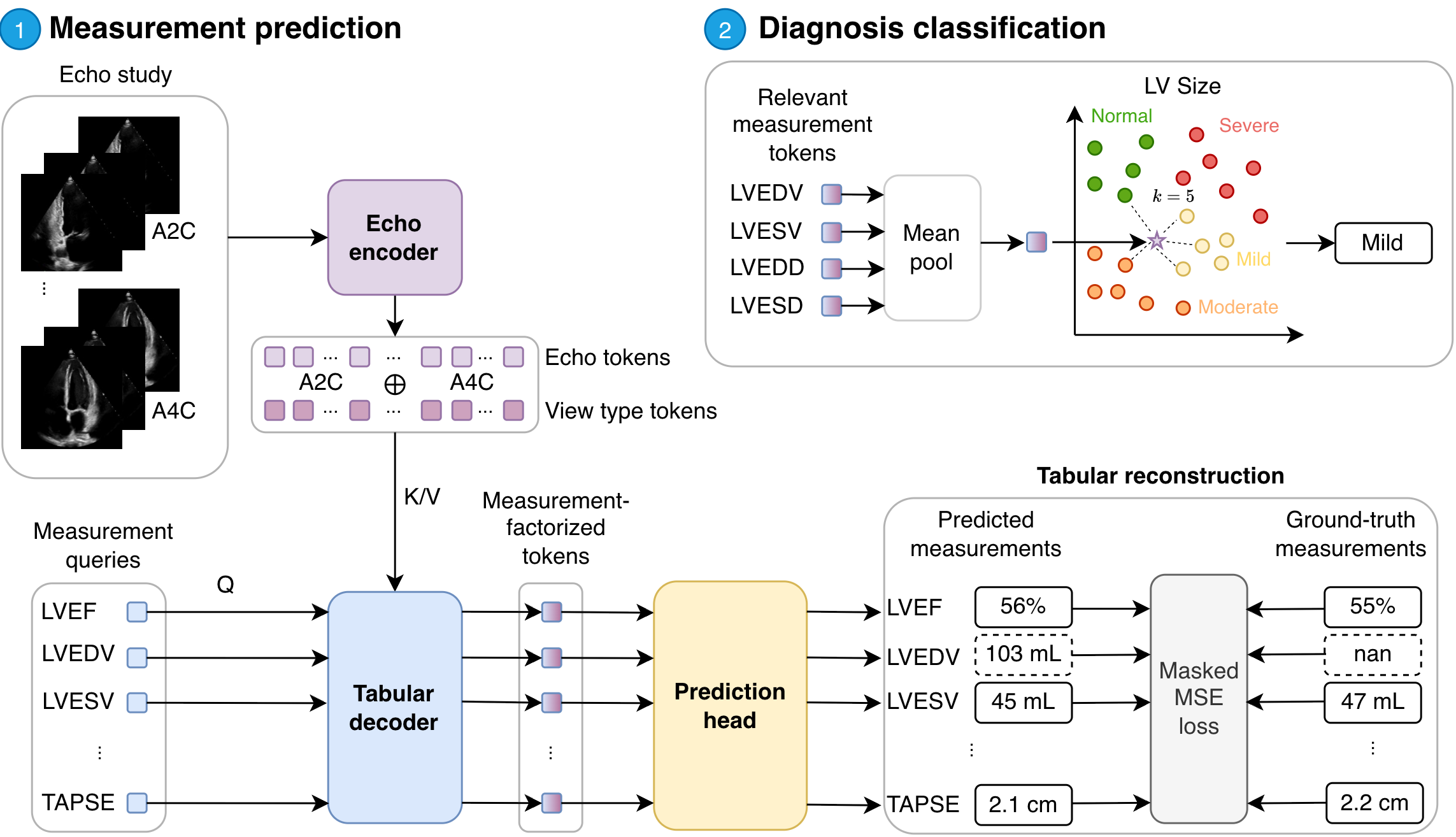}
    \caption{\textbf{Overview of \echotab{}, our measurement-grounded representation learning framework for multi-view echo studies.} (1) Given an echo study, each individual view is first encoded by the echo encoder. Next, a tabular decoder lets each measurement query selectively pull information from the echo views that are relevant for estimating that measurement. This results in a set of measurement-factorized tokens for the echo study, which the prediction head uses to predict measurement values. \echotab{} is trained with a tabular reconstruction objective, in which the MSE loss between available ground-truth and predicted measurements is calculated. (2) \echotab{}'s measurement-factorized tokens enables accurate diagnosis by using only the measurement tokens that are relevant for the downstream diagnosis.}
    \label{fig:echotab}
\end{figure}

We propose \echotab{}, a measurement-grounded representation learning framework for multi-view echo studies. As shown in \Cref{fig:echotab}, given an echo study, each individual view is first encoded by the echo encoder (\Cref{sec:echo_encoder}), resulting in a set of echo tokens per view. A tabular decoder with measurement queries (\Cref{sec:tabular_decoder}) then associates each measurement with the echo views most relevant to it, producing measurement tokens that the prediction head maps to the final measurements (\Cref{sec:prediction_head}). The framework is trained end-to-end with a tabular reconstruction objective (\Cref{sec:tabular_reconstruction}), after which relevant measurement tokens can be selected for accurate downstream diagnosis.

\subsection{Echo Encoder}
\label{sec:echo_encoder}

An echocardiography study \(S=\{v_1,\dots,v_N\}\) consists of \(N\) video clips, each assigned a view label \(\kappa(v_i)\in\mathcal{C}\), where \(\mathcal{C}\) denotes the set of view types (e.g., A4C and PLAX). Let \(\mathcal{V}\subseteq\mathcal{C}\) denote the subset of view types present in a given study. For each \(c\in\mathcal{V}\), we select exactly one video clip \(v^{(c)}\in S\) satisfying \(\kappa(v^{(c)})=c\). During training, the selected clip for each available view is sampled uniformly at random from all clips in \(S\) with label \(c\). At inference, the first such clip is selected deterministically.

The echo encoder is initialized with EchoPrime \citep{echoprime} weights and uses the MViTv2 \citep{mvitv2} architecture. Each selected view instance \(v^{(c)}\) is encoded independently, and we retain \(L_{v^{(c)}}\) local patch tokens \(\{z_{v^{(c)},j}\}_{j=1}^{L_{v^{(c)}}}\) rather than the global \texttt{CLS} token, where each \(z_{v^{(c)},j}\) represents a spatio-temporal patch of the video clip. The local tokens from all selected views are concatenated into the study-level echo token sequence,

\begin{equation}
Z = \bigl[ z_{v^{(c)},j} + e_c ,\bigr]_{c \in \mathcal{V}, j=1,\ldots,L_{v^{(c)}}},
\end{equation}

where \(e_c \in \mathbb{R}^{d}\) is a learnable view embedding for view type \(c\). In the subsequent step, we mask background patch tokens, which are those with mean pixel intensity below the threshold \(\tau=-0.6\) (corresponding to approx. \(0\) before normalization). Although MViTv2's \citep{mvitv2} multiscale architecture already endows each local token with spatial context beyond its own patch, we empirically find that masking background tokens improves view classification performance (App. \ref{app:echo_tokens}), suggesting that representations based primarily on foreground regions are more discriminative for view identification.

\subsection{Tabular decoder}
\label{sec:tabular_decoder}

Each measurement in an echo study is derived from a subset of echo views. For example, the left ventricular ejection fraction measurement is  estimated from the apical two-chamber (A2C) and four-chamber (A4C) views. We therefore design the tabular decoder to allow each measurement token to selectively aggregate information from the echo views relevant to its estimation.

The tabular decoder consists of $B$ cross-attention blocks, each followed by a feed-forward network (FFN), with residual connections, layer normalization, and learnable layer-scale parameters. We initialize learnable measurement queries $q_m \in \mathbb{R}^{d}$ for measurement $m\in{\{1,\ldots,M\}}$. These serve as queries, while the echo token sequence from \Cref{sec:echo_encoder} serve as keys and values. Before entering the decoder, the two token sets are mapped to a common dimension $d$ using separate linear projections. The tabular decoder transforms each $q_m$ into a measurement-factorized representation $x_m\in\mathbb{R}^{d}$, which is subsequently used to predict measurement $m$. We omit self-attention between measurement tokens to keep the resulting measurement representations independent of one another, yielding a factorized representation for downstream diagnosis.

\subsection{Prediction Head}
\label{sec:prediction_head}

\begin{figure}[t]
    \centering
        \centering
        \includegraphics[width=\linewidth]{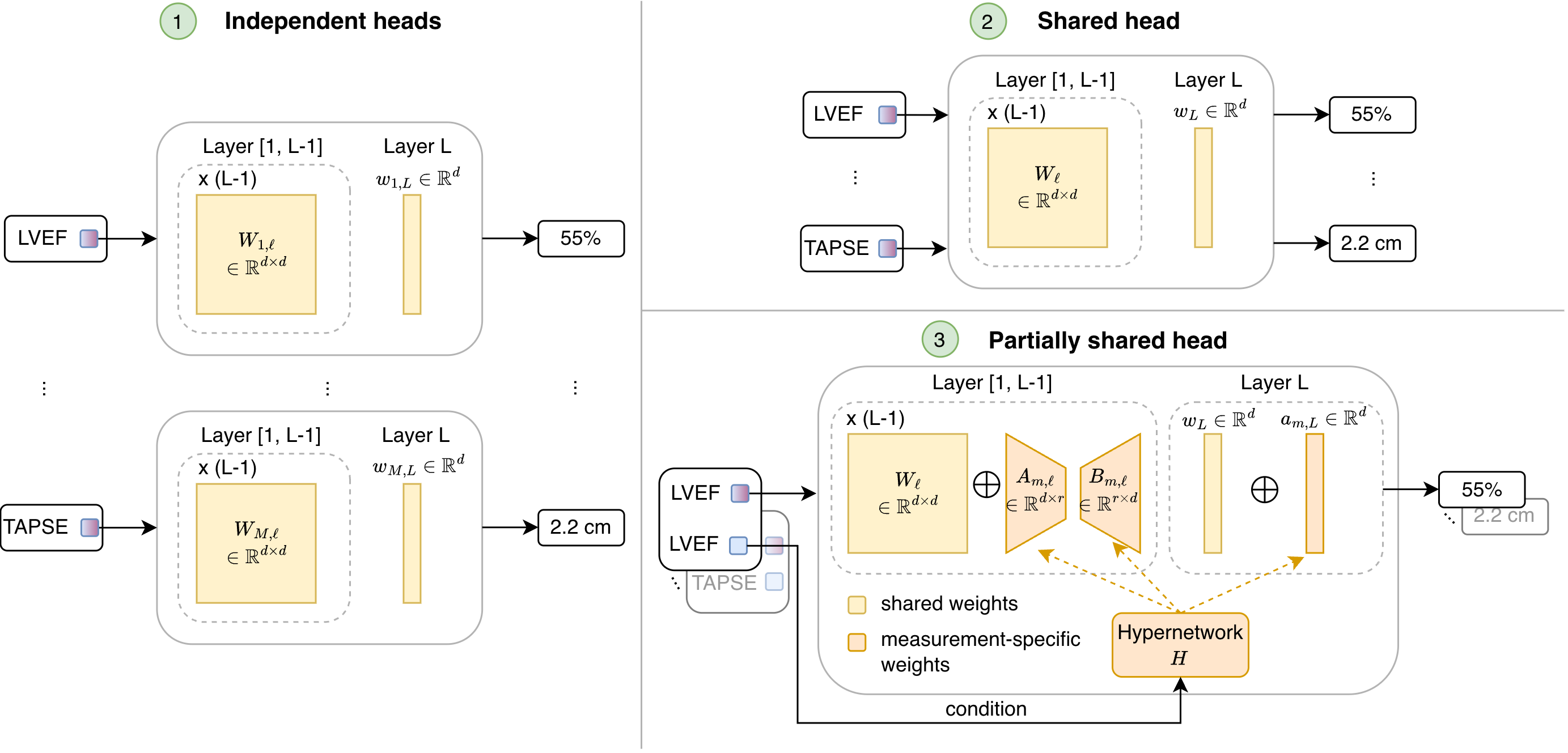}
    \caption{\textbf{Variants of the prediction head used in \echotab{}}. The prediction head is used during the tabular reconstruction of continuous measurements from their corresponding measurement-specific representations. We compare different degrees of parameter sharing to assess the trade-off between shared and measurement-specific components of the prediction function. (1) \textbf{Independent heads}: each measurement token has its own MLP head, enabling fully measurement-specific parameters. (2) \textbf{Shared head}: all measurement tokens share the same MLP head, enforcing fully shared parameters. (3) \textbf{Partially shared head}: all measurement tokens share the base weights of the MLP head, while a hypernetwork generates measurement-specific parameter updates, providing a compromise between shared and measurement-specific parameters. Biases are omitted for clarity.
    }
    \label{fig:prediction_head}
\end{figure}

The prediction head acts as a lightweight read-out on top of the tabular decoder in \Cref{sec:tabular_decoder}: given the measurement-factorized tokens from the tabular decoder, it reconstructs the corresponding tabular measurement values. As shown in \Cref{fig:prediction_head}, we propose three variants of the prediction head: independent, shared, and partially shared. Throughout, $\sigma(\cdot)$ denotes the chosen element-wise non-linearity (e.g., GELU), and $h_m^{(0)} = x_m$.

\paragraph{Independent heads.}

This is the standard multi-task setting, where each measurement token is assigned a separate $L$-layer MLP head. For measurement $m \in \{1,\dots,M\}$,

\begin{equation}
\begin{aligned}
h_m^{(\ell)} &= \sigma\!\left(W_{m,\ell} h_m^{(\ell-1)} + b_{m,\ell}\right),
\quad \ell = 1,\ldots,L-1, \\
\hat{y}_{m} &= w_{m,L}^\top h_m^{(L-1)} + b_{m,L},
\end{aligned}
\end{equation}

with $W_{m,\ell}\!\in\!\mathbb{R}^{d\times d}$, $b_{m,\ell}\!\in\!\mathbb{R}^{d}$, $w_{m,L}\!\in\!\mathbb{R}^{d}$, and $b_{m,L}\!\in\!\mathbb{R}$. Thus, each measurement learns an independent set of prediction-head parameters, and the total number of parameters scales linearly with the number of measurements, i.e., $O(M)$.

\paragraph{Shared head.}

A single $L$-layer MLP head is shared across all measurement tokens. For measurement $m \in \{1,\dots,M\}$,

\begin{equation}
\begin{aligned}
h_m^{(\ell)} &= \sigma\!\left(W_{\ell} h_m^{(\ell-1)} + b_{\ell}\right),
\quad \ell = 1,\ldots,L-1, \\
\hat{y}_{m} &= w_L^\top h_m^{(L-1)} + b_L,
\end{aligned}
\end{equation}

with $W_{\ell}\!\in\!\mathbb{R}^{d\times d}$, $b_{\ell}\!\in\!\mathbb{R}^{d}$, $w_{L}\!\in\!\mathbb{R}^{d}$, and $b_{L}\!\in\!\mathbb{R}$. The same parameters are therefore used to predict every measurement, making the number of prediction-head parameters independent of the number of measurements, i.e., $O(1)$ w.r.t. $M$.

\paragraph{Partially shared head.}

Inspired by text-to-LoRA \citep{texttolora}, which uses a hypernetwork \citep{ha2017hypernetworks} conditioned on a task description to generate task-specific low-rank corrections for an LLM, the partially shared head analogously treats the shared $L$-layer MLP as the base model and a hypernetwork conditioned on the measurement query $q_m$  generates measurement-specific low-rank corrections on top of the shared weights. For measurement $m \in \{1,\dots,M\}$,

\begin{equation}
\begin{aligned}
h_m^{(\ell)} &= \sigma\!\left(\left(W_\ell + \tfrac{\alpha}{r} A_{m,\ell} B_{m,\ell}\right) h_m^{(\ell-1)} + b_\ell + \beta_{m,\ell}\right),
\quad \ell = 1,\ldots,L-1, \\
\hat{y}_m &= \left(w_L + a_{m,L}\right)^\top h_m^{(L-1)} + b_L + \beta'_{m,L},
\end{aligned}
\end{equation}

where $W_\ell\!\in\!\mathbb{R}^{d\times d}$, $b_\ell\!\in\!\mathbb{R}^{d}$, $w_L\!\in\!\mathbb{R}^{d}$, and $b_L\!\in\!\mathbb{R}$ are shared, while $A_{m,\ell}\!\in\!\mathbb{R}^{d\times r}$, $B_{m,\ell}\!\in\!\mathbb{R}^{r\times d}$ with $r\!\ll\!d$, $\beta_{m,\ell}\!\in\!\mathbb{R}^{d}$, $a_{m,L}\!\in\!\mathbb{R}^{d}$, and $\beta'_{m,L}\!\in\!\mathbb{R}$ are measurement-specific corrections generated by a hypernetwork $H$ conditioned on $q_m$. Following \citep{hu2022lora}, the low-rank update is scaled by $\frac{\alpha}{r}$, where $r$ is the rank and $\alpha$ is a scaling hyperparameter. We investigate the impact of various hypernetwork variants in \Cref{tab:ablation_hypernetwork_variants}. 
\subsection{Tabular Reconstruction Objective}
\label{sec:tabular_reconstruction}

Echo measurements are continuous and sparse: each study reports only a subset of them. They also vary widely in unit and scale. We therefore train \echotab{} with a masked, variance-normalized mean squared error. For a mini-batch in which $\mathcal{O}$ is the set of observed (study, measurement) pairs,

\begin{equation}
    \mathcal{L} = \frac{1}{|\mathcal{O}|} \sum_{(i,m) \in \mathcal{O}}
    \frac{(\hat{y}_{i,m} - y_{i,m})^2}{s_m^2},
\end{equation}

where $\hat{y}_{i,m}$ and $y_{i,m}$ are the predicted and true values of measurement $m$
in study $i$, and $s_m^2$ is the variance of measurement $m$ over the training set. Dividing by $s_m^2$ puts all measurements on a comparable footing, so the loss is not dominated by measurements with large absolute values.

\section{Experiments}
\label{Experiments}

\subsection{Data}
\label{sec:data}

We train \echotab{} on two datasets separately: MIMIC-IV-Echo v1.0 \citep{PhysioNet-mimic-iv-echo-1.0, pollard_physionet_2026} and an internal dataset. After data processing, MIMIC-IV-Echo contains 4,514 patients with 7,144 studies and our internal cohort 7,404 patients with 9,431 studies (App. \ref{app:data_processing} \& \ref{app:measurements}).
\subsection{Baselines and Evaluation Strategy}
\label{sec:baselines}

We evaluate \echotab{} on diagnosis classification tasks on two datasets, covering conditions ranging from valvular disease to systolic dysfunction (App.~\ref{app:diagnosis_clf_dataset_details}). For each task, we train a $k$-nearest neighbors classifier with $k=5$ on the resulting study-level embeddings. We compare \echotab{} against existing echo foundation models and three classes of baselines trained on the tabular measurements available in each dataset. First, a contrastive baseline aligns paired echo and measurement representations, encoded by the echo encoder and the pre-trained TARTE tabular encoder \citep{tarte} respectively, in a shared embedding space, following the image-tabular formulation of MMCL \citep{mmcl}. Second, a supervised contrastive (SupCon) baseline uses tabular similarity to define positive and negative echo pairs, following the idea of TGV \citep{tgv} and inspired by the original SupCon \citep{khosla2020supervised}. Third, an MLP baseline adds independent measurement-prediction heads to the echo encoder and is trained using the tabular reconstruction objective of \echotab{}. Details on these baselines are provided in App. \ref{app:baselines}.

By default, all models treat an echo study as the sample unit. However, for the contrastive and SupCon baselines, we additionally include view-level training to maximize batch size. Both these baselines are trained using 8 A40 GPUs, whereas the MLP baseline and \echotab{} use a single A100 GPU with 32~GB of memory. Hyperparameters used by \echotab{} are detailed in App. \ref{app:hyperparameters}.

\echotab{} factorizes the input echo study into measurement tokens $\{x_m\}_{m=1}^{M}$. For diagnosis $k$, let $\mathcal{M}_k$ denote the measurements relevant to its prediction (see \Cref{tab:diagnosis_measurement_view_mapping_mimic} and \Cref{tab:diagnosis_measurement_view_mapping_internal}). We construct a diagnosis-specific study embedding as

\begin{equation}
s_k = \frac{1}{|\mathcal{M}_k|}\sum_{m\in\mathcal{M}_k} x_m.
\end{equation}

As the baselines do not provide this factorization into measurement-specific tokens, we instead use the study-level echo embedding obtained by averaging representations across available views.

In App.~\ref{app:view_prior_baselines}, we investigate whether the advantage of \echotab{} is solely from its diagnosis-specific prior, by giving the baselines an analogous prior in the form of diagnosis-relevant views. In App.~\ref{app:ablation_echo_views}, we examine whether \echotab{} benefits from restricting its input to these relevant views.

\subsection{Diagnosis Classification}
\label{sec:diagnosis_classification}

\begin{table}[!t]
\centering

\caption{
\textbf{\echotab{}'s measurement-specific representations enable accurate diagnosis for multi-view echocardiography.}
By enabling diagnosis-specific selection of relevant measurement tokens, \echotab{} consistently outperforms echo foundation models, as well as study- and view-level contrastive and MLP baselines, across downstream diagnosis classification tasks. Unlike contrastive approaches, \echotab{} does not require large batch sizes, enabling efficient training. Bold and underlined values denote the \textbf{best} and \underline{second-best} results.
}
\label{tab:diagnosis_classification}

\begin{subtable}{\textwidth}
\centering
\caption{Diagnosis classification results on MIMIC (AUC).}
\label{tab:diagnosis_classification_mimic}
\tiny
\setlength{\tabcolsep}{2.8pt}
\begin{tabular}{ll ll l ccc cccc cc c}
\toprule
\multirow{3}{*}{\textbf{Method}}
  & \multirow{3}{*}{\makecell[l]{\textbf{Echo}\\\textbf{encoder}}}
  & \multicolumn{2}{c}{\textbf{Echo-tabular pre-training}}
  & \multicolumn{1}{c}{\textbf{Inference config.}}
  & \multicolumn{10}{c}{\textbf{MIMIC}} \\
\cmidrule(lr){3-4} \cmidrule(lr){5-5} \cmidrule(lr){6-15}
& & \multirow{2}{*}[-0.5ex]{\makecell[l]{\textbf{Role of}\\\textbf{tabular data}}}
  & \multirow{2}{*}[-0.5ex]{\makecell[l]{\textbf{Echo}\\\textbf{sample unit}}}
  & \multirow{2}{*}[-0.5ex]{\makecell[l]{\textbf{Measurement}\\\textbf{tokens}}}
  & \multicolumn{3}{c}{\textbf{Valves}}
  & \multicolumn{4}{c}{\textbf{Size}}
  & \multicolumn{2}{c}{\textbf{Sys. function}}
  & \multirow{2}{*}{\textbf{Avg.}} \\
\cmidrule(lr){6-8} \cmidrule(lr){9-12} \cmidrule(lr){13-14}
& & & & &
\textbf{AS} & \textbf{AR} & \textbf{MR} & \textbf{LA} & \textbf{RA} & \textbf{LV} & \textbf{RV} & \textbf{LVEF} & \textbf{RVEF} & \\
\midrule

\multicolumn{15}{l}{\textit{Echo foundation models (Reference)}} \\
PanEcho & PanEcho & -- & -- & --
    & .597 & \textbf{.644} & .609 & .708 & .717 & \textbf{.877} & .646 & .775 & .748 & .702 \\
EchoJEPA-L & EchoJEPA-L & -- & -- & --
    & .525 & .589 & .557 & .628 & .730 & .644 & .660 & .630 & .721 & .632 \\
EchoPrime & EchoPrime & -- & -- & --
    & .526 & .558 & \underline{.618} & .700 & .724 & .705 & .638 & .736 & .725 & .659 \\
\midrule

\multicolumn{15}{l}{\textit{Baselines}} \\
Contrastive & EchoPrime & Tabular alignment target & view & --
    & .552 & .546 & .615 & .708 & \underline{.782} & .759 & .694 & .737 & .753 & .683 \\
Contrastive & EchoPrime & Tabular alignment target & study & --
    & .583 & .547 & \underline{.618} & .725 & \underline{.782} & .789 & .737 & .693 & .732 & .690 \\
SupCon & EchoPrime & Tabular guidance & view & --
    & \underline{.672} & .575 & .599 & \underline{.735} & .778 & .684 & .704 & \textbf{.844} & .749 & \underline{.704} \\
SupCon & EchoPrime & Tabular guidance & study & --
    & .592 & .567 & .599 & .725 & .754 & .733 & \underline{.743} & .772 & \textbf{.762} & .694 \\
MLP & EchoPrime & Tabular reconstruction & study & --
    & .642 & .551 & .573 & .713 & .774 & .745 & \underline{.743} & .775 & \underline{.759} & .697 \\
\midrule

\multicolumn{15}{l}{\textit{Ours}} \\
\echotab{} & EchoPrime & Tabular reconstruction & study & Relevant
    & \textbf{.723} & \underline{.592} & \textbf{.658} & \textbf{.745} & \textbf{.808} & \underline{.852} & \textbf{.756} & \underline{.843} & .754 & \textbf{.748} \\
\bottomrule
\end{tabular}
\end{subtable}

\vspace{1em}

\begin{subtable}{\textwidth}
\centering
\caption{Diagnosis classification results on the internal dataset (AUC).}
\label{tab:diagnosis_classification_internal}
\tiny
\setlength{\tabcolsep}{3pt}
\begin{tabular}{ll ll l cc cccc cc c}
\toprule
\multirow{3}{*}{\textbf{Method}}
  & \multirow{3}{*}{\makecell[l]{\textbf{Echo}\\\textbf{encoder}}}
  & \multicolumn{2}{c}{\textbf{Echo-tabular pre-training}}
  & \multicolumn{1}{c}{\textbf{Inference config.}}
  & \multicolumn{9}{c}{\textbf{Internal}} \\
\cmidrule(lr){3-4} \cmidrule(lr){5-5} \cmidrule(lr){6-14}
& & \multirow{2}{*}[-0.5ex]{\makecell[l]{\textbf{Role of}\\\textbf{tabular data}}}
  & \multirow{2}{*}[-0.5ex]{\makecell[l]{\textbf{Echo}\\\textbf{sample unit}}}
  & \multirow{2}{*}[-0.5ex]{\makecell[l]{\textbf{Measurement}\\\textbf{tokens}}}
  & \multicolumn{2}{c}{\textbf{Valves}}
  & \multicolumn{4}{c}{\textbf{Size}}
  & \multicolumn{2}{c}{\textbf{Sys. function}}
  & \multirow{2}{*}{\textbf{Avg.}} \\
\cmidrule(lr){6-7} \cmidrule(lr){8-11} \cmidrule(lr){12-13}
& & & & &
\textbf{AS} & \textbf{MR} & \textbf{LA} & \textbf{RA} & \textbf{LV} & \textbf{RV} & \textbf{LVEF} & \textbf{RVEF} & \\
\midrule

\multicolumn{14}{l}{\textit{Echo foundation models (Reference)}} \\
PanEcho & PanEcho & -- & -- & --
    & .681 & .685 & .698 & .750 & .792 & .672 & .741 & .593 & .702 \\
EchoJEPA-L & EchoJEPA-L & -- & -- & --
    & .606 & .641 & .682 & .735 & .600 & .565 & .666 & .656 & .644 \\
EchoPrime & EchoPrime & -- & -- & --
    & .740 & .711 & .721 & \textbf{.790} & .772 & .660 & .765 & .684 & .730 \\
\midrule

\multicolumn{14}{l}{\textit{Baselines}} \\
Contrastive & EchoPrime & Tabular alignment target & view & --
    & .734 & .684 & .760 & .777 & .791 & \underline{.747} & .762 & .727 & .748 \\
Contrastive & EchoPrime & Tabular alignment target & study & --
    & .785 & .696 & .771 & .777 & .807 & .697 & .756 & .692 & .748 \\
SupCon & EchoPrime & Tabular guidance & view & --
    & .777 & .669 & .779 & \underline{.786} & \underline{.816} & .739 & .801 & .692 & .757 \\
SupCon & EchoPrime & Tabular guidance & study & --
    & \textbf{.809} & .696 & \textbf{.792} & .760 & \textbf{.830} & .703 & .785 & .627 & .750 \\
MLP & EchoPrime & Tabular reconstruction & study & --
    & .790 & \textbf{.717} & .775 & .765 & .815 & .685 & \underline{.810} & \underline{.731} & \underline{.761} \\
\midrule

\multicolumn{14}{l}{\textit{Ours}} \\
\echotab{} & EchoPrime & Tabular reconstruction & study & Relevant
    & \underline{.800} & \underline{.713} & \underline{.787} & .766 & \textbf{.830} & \textbf{.749} & \textbf{.829} & \textbf{.749} & \textbf{.778} \\
\bottomrule
\end{tabular}
\end{subtable}

\end{table}

\Cref{tab:diagnosis_classification} presents diagnosis classification results on MIMIC and our internal dataset. \echotab{} consistently improves diagnosis classification across both datasets and achieves the highest average AUC among echo foundation models, contrastive baselines, and the MLP baseline. It ranks first or second on 8 of 9 diagnoses on MIMIC, and on 7 of 8 diagnoses on our internal dataset. Relative to EchoPrime (the shared initialization used by all baselines), \echotab{} improves AUC on all 9 MIMIC diagnoses, increasing the average AUC from 0.659 to 0.748 (+13.5\%). On the internal dataset, it improves performance on 7 of 8 diagnoses and raises the average AUC from 0.730 to 0.778 (+6.6\%). All baselines also improve upon EchoPrime in average AUC, indicating that echo-tabular pre-training improves diagnosis accuracy.

On MIMIC, \echotab{} outperforms the strongest baseline (view-level SupCon) by 4.4 average AUC points (0.748 vs. 0.704), while requiring only 1 GPU compared with 8 GPUs for SupCon. On the internal dataset, it outperforms the strongest baseline (MLP) on 7 of 8 diagnoses and improves average AUC by 1.7 points (0.778 vs. 0.761), despite using the same tabular reconstruction objective. This suggests that \echotab{}'s gains are not attributable only to the training objective, but also to its usage of diagnosis-relevant measurement tokens for accurate prediction (\Cref{sec:architectural_ablation} for more).

\subsection{Attention Map Analysis}
\label{sec:attention_map_analysis}

\begin{figure}[t]
    \centering
        \centering
        \includegraphics[width=0.85\linewidth]{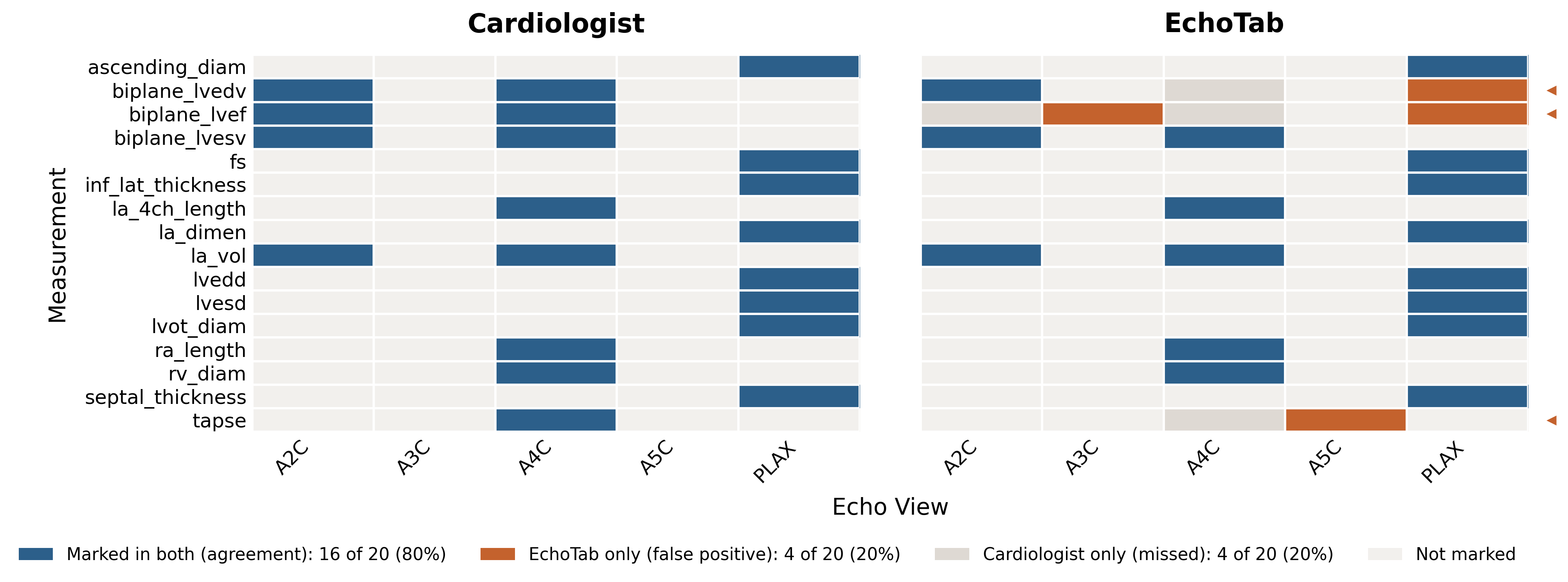}
    \caption{\textbf{Comparison of the echo views a cardiologist (left) and \echotab{} (right) rely on when deriving measurements on MIMIC.} \echotab{} matches the cardiologist’s view in 16 of 20 (80\%) measurement-view pairs, indicating it learns clinically meaningful measurement-view associations.}
    \label{fig:cross_attention}
\end{figure}

\echotab{}'s tabular decoder allows us to examine which echo views it attends to for individual measurements through its cross-attention maps. \Cref{fig:cross_attention} compares its attention pattern with cardiologist-identified relevant views on MIMIC. We first average the attention maps across all cross-attention layers and then binarize the resulting map to match the number of views deemed relevant by the cardiologist: if $x$ views are considered relevant, the $x$ highest-attended views are labeled relevant. Measurements that depend on view types absent from the view set $\mathcal{C}$ or on Doppler views are excluded. The latter are omitted because the view classifier groups all Apical Doppler acquisitions into a single broad category, which is too coarse for this measurement-view analysis.

\echotab{} selects the same echo views as the cardiologist in 16 of 20 (80\%) measurement-view pairs, agreeing on 13 of 16 (81\%) measurements. The three disagreements involve biplane LVEDV and LVEF (left ventricle measurements) and TAPSE (right ventricular function). For TAPSE, it favors A5C over A4C. For LVEDV and LVEF, it partly shifts attention from A2C/A4C to PLAX/A3C. In each case, the alternative views \echotab{} attends to also capture the relevant cardiac structure. Because echo views contain partially redundant information, the model may exploit correlated signals rather than strictly follow clinical protocols. Overall, \echotab{}'s attention captures clinically meaningful measurement-view associations, with deviations remaining anatomically plausible.

\section{Ablations}
\label{ablations}

\subsection{Architectural Ablation}
\label{sec:architectural_ablation}

In \Cref{tab:ablation_architecture}, we ablate the individual components of \echotab{}. Specifically, we investigate the effect of the tabular decoder, compare measurement queries with a single query shared across all measurements, and evaluate the prediction head variants described in \Cref{sec:prediction_head}. When present, the tabular decoder uses two cross-attention layers, and all prediction heads are two-layer MLPs. For experiments using measurement queries, we use diagnosis-relevant measurement tokens for diagnosis classification.

With independent heads, adding the tabular decoder improves diagnosis classification over the MLP baseline on both MIMIC (0.697 vs. 0.720) and the internal cohort (0.761 vs. 0.766). This suggests that querying echo views through the tabular decoder produces a study-level representation that is more informative for diagnosis classification than simply averaging the echo-view embeddings.

When prediction heads are independent, measurement queries slightly degrade performance relative to a single query (0.709 vs. 0.720 on MIMIC; 0.764 vs. 0.766 on internal). We attribute this to a trade-off between measurement-specific representational flexibility and cross-measurement representation alignment: separate queries allow measurements to be represented independently, but independent heads do not enforce a shared representation space among the resulting tokens. Hence, averaging these tokens can produce a less coherent study embedding for diagnosis classification.

A shared prediction head addresses this issue by mapping all measurement tokens into a common value space, while still preserving their distinct query identities. This retains measurement-specific representations while restoring cross-measurement representation alignment, making their average more coherent. Diagnosis classification consequently improves (0.732 on MIMIC; 0.773 on internal). Without measurement-specific queries, a shared head yields degenerate results because the same representation is used for every measurement, preventing measurement-specific predictions.

The partially shared head provides a compromise between measurement-specific representational flexibility and cross-measurement alignment, yielding the best diagnosis classification in both cohorts (0.748 on MIMIC; 0.778 on internal), corresponding to relative gains of 7.3\% and 2.2\% over the MLP baseline. It also achieves the best measurement prediction on MIMIC (0.418) and is essentially tied on the internal cohort (0.445) with the fully shared head (0.447) and the MLP (0.446).

\begin{table}[htbp]
\centering
\scriptsize
\setlength{\tabcolsep}{4pt}
\begin{tabular}{l cc ll cc cc}
\toprule
\multirow{2}{*}{\textbf{Method}}
  & \multirow{2}{*}{\makecell{\textbf{Echo}\\\textbf{encoder}}}
  & \multirow{2}{*}{\makecell{\textbf{Tabular}\\\textbf{decoder}}}
  & \multirow{2}{*}{\makecell[l]{\textbf{Tabular decoder}\\\textbf{queries}}}
  & \multirow{2}{*}{\makecell[l]{\textbf{Prediction}\\\textbf{head}}}
  & \multicolumn{2}{c}{\textbf{MIMIC}} & \multicolumn{2}{c}{\textbf{Internal}} \\
\cmidrule(lr){6-7} \cmidrule(lr){8-9}
& & & & & \textbf{Diag. clf.} & \textbf{Meas. pred.} & \textbf{Diag. clf.} & \textbf{Meas. pred.} \\
\midrule

MLP & \multirow{2}{*}{\cmark} & \multirow{2}{*}{\xmark} & \multirow{2}{*}{--}
    & Independent
    & .697 & .392
    & .761 & \underline{.446} \\
& & & & Shared
    & .683 & -16.944
    & .648 & -14.799 \\
\midrule

& \multirow{2}{*}{\cmark} & \multirow{2}{*}{\cmark} & \multirow{2}{*}{Single}
    & Independent
    & .720 & .380 & .766 & .422 \\
& & & & Shared
    & .628 & -16.987
    & .624 & -14.902 \\
\midrule

\multirow{3}{*}{\echotab{}} & \multirow{3}{*}{\cmark} & \multirow{3}{*}{\cmark} & \multirow{3}{*}{Measurement}
    & Independent
    & .709 & .401 & .764 & .415 \\
& & & & Shared
    & \underline{.732} & \underline{.412} & \underline{.773} & \textbf{.447} \\
& & & & Partially shared
    & \textbf{.748} & \textbf{.418} & \textbf{.778} & .445 \\
\bottomrule
\end{tabular}
\caption{\textbf{Ablation of tabular decoder, queries, and prediction head in \echotab{}.} Bold and underlined values denote the \textbf{best} and \underline{second-best} results. Diagnosis classification is measured by AUC averaged across diagnoses and measurement prediction by $R^2$ averaged across measurements. In \echotab{}, a partially shared head makes the resulting measurement tokens most effective, balancing cross-measurement representation alignment with measurement-specific representational flexibility. As a result, it achieves the best diagnosis performance on both datasets.}
\label{tab:ablation_architecture}
\end{table}

\subsection{Measurement Tokens for Diagnosis Classification}
\label{sec:measurement_tokens}

In \Cref{tab:ablation_measurement_tokens}, we ablate measurement-token configurations for diagnosis classification. In the \textit{Relevant} setting, we average diagnosis-relevant measurement tokens. In \textit{All}, we average all measurement tokens available during training. As an \textit{Incorrect} control, we replace each diagnosis-specific subset with an equal-sized set of measurements sampled uniformly at random (without replacement) from other diagnoses, excluding that diagnosis’s measurements. All settings use a partially shared head.

\begin{wraptable}{r}{0.5\textwidth}
\centering
\vspace{-1em}
{\tiny
\setlength{\tabcolsep}{4pt}
\begin{tabular}{l cc}
\toprule
\textbf{Measurement tokens} & \textbf{MIMIC} & \textbf{Internal} \\
\midrule
Diagnosis-relevant & \textbf{.748} & \textbf{.778} \\
All                & .693          & .751          \\
Incorrect          & .687          & .757          \\
\bottomrule
\end{tabular}}
\caption{\textbf{Ablation of the measurement tokens used for diagnosis classification} (avg. AUC). Choosing diagnosis-relevant measurement tokens is essential for accurate diagnosis.}
\label{tab:ablation_measurement_tokens}
\end{wraptable}

Averaging all measurement tokens mixes in diagnosis-irrelevant information and reduces classification performance. Likewise, using incorrect tokens also degrades performance. These results indicate that the learned measurement tokens encode measurement-specific information and that choosing diagnosis-relevant measurement tokens is crucial for accurate diagnosis.

\section{Related Work}
\label{related_work}

\noindent\textbf{Echo foundation models.} To our knowledge, no existing echo foundation model explicitly incorporates measurements as an intermediate step for diagnosis during training. EchoJEPA \citep{echojepa} and Echo-Vision-FM \citep{echovisionfm} use self-supervised masked latent and masked pixel prediction objectives respectively, without measurements. EchoPrime \citep{echoprime} incorporates measurements only indirectly via video-text contrastive learning. In this setting, measurements are encoded as discrete text tokens, and diagnosis relies on retrieving semantically similar reports rather than reasoning from measurements. PanEcho \citep{panecho} jointly predicts measurements and diagnoses, but optimizes the two tasks in parallel rather than using measurements to derive diagnosis. In contrast, \echotab{} explicitly uses measurement representations as an intermediate step between echo views and diagnosis.

\noindent\textbf{Vision-tabular learning.} Building on CLIP’s image-text contrastive learning \citep{clip}, MMCL \citep{mmcl} aligns paired image and tabular embeddings in a shared latent space, encouraging the image encoder to capture information reflected in the associated tabular features. TGV \citep{tgv} instead uses tabular data to form positive and negative image pairs, encouraging the image encoder to learn representations that reflect tabular similarity. Both methods rely on contrastive objectives, which generally benefit from a larger and more diverse set of negative samples, obtained through larger batch sizes. In contrast, \echotab{} uses a tabular reconstruction objective that requires no negative samples and therefore does not rely on large batches.


\noindent\textbf{Concept bottlenecks and query-based decoders.} Concept bottleneck models \citep{koh2020concept} and their variants \citep{zarlenga2022concept, oikarinen2023labelfree, yuksekgonul2023posthoc} predict a set of semantically meaningful concepts from the input and then predict the target from those concepts, yielding an interpretable intermediate representation. Similar to concepts, the queries in query-based decoders extract information from an input, usually through attention layers. For example, object queries in DETR~\citep{carion2020detr}, output queries in Perceiver IO~\citep{jaegle2022perceiver}, and the learned queries of the Q-Former in BLIP-2~\citep{li2023blip2}. \echotab{} combines these two ideas. Like a concept bottleneck, \echotab{} uses measurement tokens as concepts that represent multi-view echos, which are subsequently used to predict diagnoses. Like a query-based decoder, it reads out these tokens from multi-view echos through cross-attention layers.

\section{Discussion and Limitations}
\label{conclusion}

\paragraph{Discussion.}

Our experiments demonstrate the value of using measurement tokens as an intermediate representation between echo views and diagnosis. This formulation factorizes a multi-view echo study into measurement-specific tokens, allowing each diagnosis to use the measurement tokens most relevant to its prediction. Using these diagnosis-relevant measurement tokens improves diagnostic accuracy compared with generic study-level representations produced by alternative pre-training methods. Moreover, because these representations are grounded in measurement values, they provide a more interpretable basis for diagnosis than a direct mapping from echo views to diagnosis.

While our study focuses on echocardiography, our representation learning framework is, in principle, modality-agnostic and applicable to multi-view imaging studies with associated  measurements. Future work could investigate its application to other modalities, such as cardiac MRI, CT, or retinal imaging, by learning measurement-specific representations from domain-specific quantitative measurements and leveraging these representations for downstream tasks. 

\paragraph{Limitations.}

\echotab{} requires specifying the mapping between diagnoses and the measurements relevant to them when constructing diagnosis-specific embeddings. While this mirrors clinical reasoning, it introduces domain knowledge into the model and requires specifying the relevant measurements when extending the framework to new diagnoses. In addition, we explore only a single strategy for combining the measurement-specific tokens relevant to each diagnosis, namely averaging them into a diagnosis-specific study embedding. More sophisticated aggregation mechanisms, such as learned weighting or attention, may better capture interactions among measurement tokens. Future work could therefore explore learning the measurement-diagnosis relationships, developing better aggregation strategies, and extending the framework to additional modalities and tasks.

\subsection*{AI use statement}

In this work, we used generative AI tools to design and provide feedback on our research methodology and experiments, implement methods, and interpret results. We have not used generative AI tools to generate synthetic datasets, help develop theoretical models or conceptual frameworks, propose or refine hypotheses, or clean and reformat datasets, and the remaining required disclosure tasks (formulating mathematical claims, providing critical ingredients for proving mathematical claims, assisting in the writing of proofs, assisting with translation, and supporting qualitative and thematic data analysis) are not applicable to this work. Additionally, we used generative AI tools to create and edit software code, draft parts of a research paper, edit the paper to improve readability, and identify relevant literature. We have reviewed all AI-assisted work. In particular, AI feedback on our research methodology and experimental design was critically assessed by the authors, and final design decisions were made by the authors. AI-assisted implementations and code were verified and tested for correctness. AI-assisted interpretations of results were used as feedback rather than the final interpretation. All literature identified with AI assistance was read and verified by the authors to confirm its existence and relevance. All AI-drafted or AI-edited text was revised by the authors to ensure it accurately reflects our work. We take responsibility for the final content of this work, including text, claims, or artifacts produced with the aid of generative AI.

\subsection*{Reproducibility statement}

We provide our code at \url{https://github.com/michelleespranita/EchoTab}, which includes the implementation of our model architecture and the downstream task. Our experiments use the MIMIC-IV-Echo dataset v1.0 \citep{PhysioNet-mimic-iv-echo-1.0, pollard_physionet_2026}, which is distributed through PhysioNet under credentialed access. Researchers can obtain access by completing PhysioNet's credentialing process. We describe our data processing pipeline in Appendix \ref{app:data_processing} and the list of measurements we use in Appendix \ref{app:measurements}. We report our model's hyperparameters in Appendix \ref{app:hyperparameters} and further details on the baselines in Appendix \ref{app:baselines}.




\subsubsection*{Acknowledgments}
This study was approved by the Ethics Committee of TUM Klinikum Rechts der Isar (reference number 2025-395-S-CB, application dated July 13, 2025).

\bibliography{iclr2027_conference}
\bibliographystyle{iclr2027_conference}

\newpage
\appendix
\label{Appendix}

\section*{Appendix}
\section{Data Processing}
\label{app:data_processing}

We construct paired echocardiogram-measurements datasets from two sources: an internal cohort and MIMIC-IV-Echo v1.0 \citep{PhysioNet-mimic-iv-echo-1.0, pollard_physionet_2026}. For both, measurement selection and study filtering were performed independently, and patient-level splits were used throughout so that no patient appears in more than one split.

\paragraph{Measurement selection.}
Starting from all measurements available, we applied the filtering steps summarized in Table~\ref{tab:measurement_selection}.

\begin{table}[h]
  \centering
  \footnotesize
  \caption{Number of measurements remaining after each filtering step. Dashes indicate
  steps that do not apply to the given dataset.}
  \label{tab:measurement_selection}
  \begin{tabular}{lcc}
    \toprule
    \textbf{Step} & \textbf{Internal} & \textbf{MIMIC} \\
    \midrule
    Initial measurements                                      & 647 & 302 \\
    Remove all-missing measurements                           & 221 & 300 \\
    Remove stress-echo measurements                           & --  & 233 \\
    Remove non-measurement fields (demographics, hemodynamics)& --  & 195 \\
    Remove diagnostic labels (to be used for downstream tasks)            & --  & \phantom{0}99 \\
    Remove all-missing measurements (second pass)             & --  & \phantom{0}71 \\
    Remove measurements present in ${<}100$ studies           & 121 & \phantom{0}47 \\
    Remove redundant / merge synonymous measurements          & \phantom{0}67 & \phantom{0}45 \\
    \midrule
    \textbf{Final}                                            & \textbf{67} & \textbf{45} \\
    \bottomrule
  \end{tabular}
\end{table}

For our internal cohort, redundancy removal collapsed measurements of the same quantity obtained by different acquisition or computation methods, retaining a single preferred variant per
quantity (preference order: biplane mode $>$ 4-chamber mode $>$ 2-chamber mode $>$ M-mode; 2D $>$ M-mode). We also excluded body-size-indexed measurements (e.g.\ LV mass index, stroke volume index), as these require anthropometric data that is not consistently available. For MIMIC, synonymous fields reporting identical quantities were merged (e.g., \texttt{biplane\_lvef} and \texttt{lvef}). Measurements requiring M-mode acquisition were retained in both datasets.

\paragraph{Study filtering.}

Linking structured measurements to echo studies, excluding studies with no recorded values in the retained measurement list, and excluding studies without videos from the view set $\mathcal{C}$ (A2C, A3C, A4C, A5C, PLAX, PSAX, and Apical Doppler) yields \textbf{9,430 studies} for our internal cohort and \textbf{7,063 studies} for MIMIC.

\paragraph{Measurement pre-processing.}

Within each dataset, measurement values below the 1\textsuperscript{st} or above the 99\textsuperscript{th} percentile of the corresponding training-set distribution are set to missing. Measurement units are harmonized across datasets (e.g., milliseconds to seconds for deceleration and pressure half-times, and millimeters to centimeters for TAPSE).

\paragraph{Video sampling during training.}
Each study contributes at most one video per view type from the view set $\mathcal{C}$ (A2C, A3C, A4C, A5C, PLAX, PSAX, and Apical Doppler). During training, one video is randomly sampled from the available videos for each view type at every epoch, yielding a different combination of videos across epochs. At evaluation time, sampling is replaced by the first available video of each view type to ensure deterministic evaluation. For $k$-NN classification, the first available video of each view type is used regardless of training or evaluation mode.

\paragraph{Splits.}
Both datasets are split at the patient level into training, validation, and test sets (Table~\ref{tab:splits}).

\begin{table}[h]
  \centering
  \footnotesize
  \caption{Dataset splits, reported as number of patients and number of echo studies.}
  \label{tab:splits}
  \begin{tabular}{lcc}
    \toprule
    & \textbf{Internal (patients / studies)}
    & \textbf{MIMIC (patients / studies)} \\
    \midrule
    Train
    & 6,272 / 7{,}690
    & 3,612 / 5,610 \\
    Validation
    & 565 / \phantom{0}879
    & 450 / \phantom{0}723 \\
    Test
    & 566 / \phantom{0}861
    & 452 / \phantom{0}730 \\
    \midrule
    \textbf{Total}
    & \textbf{7,403} / \textbf{9{,}430}
    & \textbf{4,514} / \textbf{7{,}063} \\
    \bottomrule
  \end{tabular}
\end{table}

\section{Additional Results on Diagnosis Classification}

\subsection{Can a diagnosis-specific prior for the baselines close the gap to \echotab{}?}
\label{app:view_prior_baselines}

\echotab{} conditions its diagnosis predictions on diagnosis-relevant measurement tokens, and this conditioning is central to its performance: using all measurement tokens instead of diagnosis-relevant ones reduces its average AUC from 0.748 to 0.693 on MIMIC and from 0.778 to 0.751 on the internal dataset (\Cref{tab:ablation_measurement_tokens}). This raises the question of whether the advantage of \echotab{} over the baselines stems solely from access to a diagnosis-specific prior. Since each measurement is acquired from specific echo views, the closest analogous prior available to the baselines is to restrict their input to the views relevant to each diagnosis. We therefore re-evaluate all baselines and echo foundation models using only diagnosis-relevant views (see \Cref{tab:diagnosis_measurement_view_mapping_mimic} and \Cref{tab:diagnosis_measurement_view_mapping_internal}). To ensure that both sides receive exactly one diagnosis-specific prior, \echotab{} still uses all views, together with its diagnosis-relevant measurement tokens.

\Cref{tab:diagnosis_classification_relevant_views} shows that incorporating diagnosis-specific priors into the baselines via diagnosis-relevant views does not close the gap: \echotab{} still achieves the highest average AUC on both datasets, outperforming the strongest baseline by 3.1 average AUC points on MIMIC (0.748 vs.\ 0.717 for MLP) and by 3.0 on the internal dataset (0.778 vs.\ 0.748 for study-level SupCon). This advantage is consistent across diagnoses: \echotab{} achieves the best average AUC on 6 of 9 diagnoses on MIMIC, and on 7 of 8 diagnoses on the internal dataset. Together, this shows that the baselines cannot replicate the advantage of \echotab{}, even when given an analogous prior through diagnosis-relevant views.

\begin{table}[!t]
\centering

\caption{
\textbf{\echotab{} still outperforms baselines when both are given diagnosis-specific prior.} Baselines and reference foundation models use only diagnosis-relevant views, whereas \echotab{} uses all views with diagnosis-relevant measurement tokens. Bold and underlined values denote the \textbf{best} and \underline{second-best} results.
}
\label{tab:diagnosis_classification_relevant_views}

\begin{subtable}{\textwidth}
\centering
\caption{Diagnosis classification results on MIMIC (AUC).}
\label{tab:diagnosis_classification_relevant_views_mimic}
\tiny
\setlength{\tabcolsep}{2.3pt}
\begin{tabular}{ll ll ll ccc cccc cc c}
\toprule
\multirow{3}{*}{\textbf{Method}}
  & \multirow{3}{*}{\makecell[l]{\textbf{Echo}\\\textbf{encoder}}}
  & \multicolumn{2}{c}{\textbf{Echo-tabular pre-training}}
  & \multicolumn{2}{c}{\textbf{Inference config.}}
  & \multicolumn{10}{c}{\textbf{MIMIC}} \\
\cmidrule(lr){3-4} \cmidrule(lr){5-6} \cmidrule(lr){7-16}
& & \multirow{2}{*}[-0.5ex]{\makecell[l]{\textbf{Role of}\\\textbf{tabular data}}}
  & \multirow{2}{*}[-0.5ex]{\makecell[l]{\textbf{Echo}\\\textbf{sample unit}}}
  & \multirow{2}{*}[-0.5ex]{\textbf{Views}}
  & \multirow{2}{*}[-0.5ex]{\makecell[l]{\textbf{Measurement}\\\textbf{tokens}}}
  & \multicolumn{3}{c}{\textbf{Valves}}
  & \multicolumn{4}{c}{\textbf{Heart dimension}}
  & \multicolumn{2}{c}{\textbf{Sys. function}}
  & \multirow{2}{*}{\textbf{Avg.}} \\
\cmidrule(lr){7-9} \cmidrule(lr){10-13} \cmidrule(lr){14-15}
& & & & & &
\textbf{AS} & \textbf{AR} & \textbf{MR} & \textbf{LA} & \textbf{RA} & \textbf{LV} & \textbf{RV} & \textbf{LVEF} & \textbf{RVEF} & \\
\midrule

\multicolumn{16}{l}{\textit{Echo foundation models (Reference)}} \\
PanEcho & PanEcho & -- & -- & Relevant & --
    & .561 & \textbf{.643} & .615 & .685 & .693 & \underline{.818} & .673 & .696 & .720 & .678 \\
EchoJEPA-L & EchoJEPA-L & -- & -- & Relevant & --
    & .499 & .490 & .506 & .615 & .710 & .656 & .679 & .609 & .697 & .607 \\
EchoPrime & EchoPrime & -- & -- & Relevant & --
    & .552 & .511 & .583 & .695 & .718 & .749 & .667 & .710 & .742 & .659 \\
\midrule

\multicolumn{16}{l}{\textit{Baselines}} \\
Contrastive & EchoPrime & Tabular alignment target & view & Relevant & --
    & .603 & .578 & .627 & \textbf{.748} & .755 & .783 & .684 & .709 & .752 & .693 \\
Contrastive & EchoPrime & Tabular alignment target & study & Relevant & --
    & \underline{.671} & .568 & .612 & .717 & \underline{.762} & .796 & \underline{.722} & .728 & \underline{.768} & .705 \\
SupCon & EchoPrime & Tabular guidance & view & Relevant & --
    & .664 & .548 & .630 & .731 & .743 & .720 & .690 & \underline{.812} & \textbf{.771} & .701 \\
SupCon & EchoPrime & Tabular guidance & study & Relevant & --
    & .635 & .514 & \underline{.652} & .719 & .709 & .746 & .714 & .750 & .745 & .687 \\
MLP & EchoPrime & Tabular reconstruction & study & Relevant & --
    & .664 & .591 & .614 & .734 & .721 & .797 & .719 & \textbf{.843} & \textbf{.771} & \underline{.717} \\
\midrule

\multicolumn{16}{l}{\textit{Ours}} \\
\echotab{} & EchoPrime & Tabular reconstruction & study & All & Relevant
    & \textbf{.723} & \underline{.592} & \textbf{.658} & \underline{.745} & \textbf{.808} & \textbf{.852} & \textbf{.756} & \textbf{.843} & .754 & \textbf{.748} \\
\bottomrule
\end{tabular}
\end{subtable}

\vspace{1em}

\begin{subtable}{\textwidth}
\centering
\caption{Diagnosis classification results on the internal dataset (AUC).}
\label{tab:diagnosis_classification_relevant_views_internal}
\tiny
\setlength{\tabcolsep}{2.8pt}
\begin{tabular}{ll ll ll cc cccc cc c}
\toprule
\multirow{3}{*}{\textbf{Method}}
  & \multirow{3}{*}{\makecell[l]{\textbf{Echo}\\\textbf{encoder}}}
  & \multicolumn{2}{c}{\textbf{Echo-tabular pre-training}}
  & \multicolumn{2}{c}{\textbf{Inference config.}}
  & \multicolumn{9}{c}{\textbf{Internal}} \\
\cmidrule(lr){3-4} \cmidrule(lr){5-6} \cmidrule(lr){7-15}
& & \multirow{2}{*}[-0.5ex]{\makecell[l]{\textbf{Role of}\\\textbf{tabular data}}}
  & \multirow{2}{*}[-0.5ex]{\makecell[l]{\textbf{Echo}\\\textbf{sample unit}}}
  & \multirow{2}{*}[-0.5ex]{\textbf{Views}}
  & \multirow{2}{*}[-0.5ex]{\makecell[l]{\textbf{Measurement}\\\textbf{tokens}}}
  & \multicolumn{2}{c}{\textbf{Valves}}
  & \multicolumn{4}{c}{\textbf{Heart dimension}}
  & \multicolumn{2}{c}{\textbf{Sys. function}}
  & \multirow{2}{*}{\textbf{Avg.}} \\
\cmidrule(lr){7-8} \cmidrule(lr){9-12} \cmidrule(lr){13-14}
& & & & & &
\textbf{AS} & \textbf{MR} & \textbf{LA} & \textbf{RA} & \textbf{LV} & \textbf{RV} & \textbf{LVEF} & \textbf{RVEF} & \\
\midrule

\multicolumn{15}{l}{\textit{Echo foundation models (Reference)}} \\
PanEcho & PanEcho & -- & -- & Relevant & --
    & .681 & .694 & .706 & .724 & .786 & .664 & .724 & .618 & .700 \\
EchoJEPA-L & EchoJEPA-L & -- & -- & Relevant & --
    & .612 & .593 & .674 & .662 & .563 & .615 & .670 & \underline{.744} & .642 \\
EchoPrime & EchoPrime & -- & -- & Relevant & --
    & .753 & .681 & .704 & \underline{.767} & .770 & .626 & .748 & .656 & .713 \\
\midrule

\multicolumn{15}{l}{\textit{Baselines}} \\
Contrastive & EchoPrime & Tabular alignment target & view & Relevant & --
    & .762 & .667 & .752 & .759 & .774 & .679 & .766 & .661 & .728 \\
Contrastive & EchoPrime & Tabular alignment target & study & Relevant & --
    & .790 & .692 & .750 & .739 & .787 & .651 & .757 & .644 & .726 \\
SupCon & EchoPrime & Tabular guidance & view & Relevant & --
    & .793 & .629 & .763 & \textbf{.769} & .764 & .719 & .790 & .685 & .739 \\
SupCon & EchoPrime & Tabular guidance & study & Relevant & --
    & \underline{.797} & .705 & \underline{.784} & .746 & \underline{.829} & \underline{.734} & .786 & .606 & \underline{.748} \\
MLP & EchoPrime & Tabular reconstruction & study & Relevant & --
    & .776 & \underline{.708} & .783 & .727 & .799 & .676 & \underline{.813} & .689 & .746 \\
\midrule

\multicolumn{15}{l}{\textit{Ours}} \\
\echotab{} & EchoPrime & Tabular reconstruction & study & All & Relevant
    & \textbf{.800} & \textbf{.713} & \textbf{.787} & .766 & \textbf{.830} & \textbf{.749} & \textbf{.829} & \textbf{.749} & \textbf{.778} \\
\bottomrule
\end{tabular}
\end{subtable}
\end{table}

\subsection{Does \echotab{} benefit from diagnosis-relevant views as prior?}
\label{app:ablation_echo_views}

\begin{wraptable}{r}{0.5\textwidth}
\centering
{\scriptsize
\setlength{\tabcolsep}{4pt}
\begin{tabular}{l l c c}
\toprule
\textbf{Views} & \textbf{Measurement tokens} & \textbf{MIMIC} & \textbf{Internal} \\
\midrule
All      & Diagnosis-relevant & \textbf{.748} & \textbf{.778} \\
Diagnosis-relevant & Diagnosis-relevant & .725 & .753 \\
\bottomrule
\end{tabular}}
\caption{\textbf{Ablation of the echo views used by \echotab{} for diagnosis classification} (average AUC). Restricting \echotab{} to diagnosis-relevant views lowers performance on both datasets, indicating that the tabular decoder benefits from access to all views rather than a defined set of relevant views.}
\label{tab:ablation_echo_views}
\end{wraptable}

We next ask whether \echotab{} itself benefits from diagnosis-specific views as prior. To this end, we restrict \echotab{} to the same diagnosis-relevant views used for the baselines in \Cref{tab:diagnosis_classification_relevant_views}, while keeping its diagnosis-relevant measurement tokens. As shown in \Cref{tab:ablation_echo_views}, this lowers the average AUC from 0.748 to 0.725 on MIMIC and from 0.778 to 0.753 on the internal dataset. This suggests that the view-measurement relationships learned by the tabular decoder are more effective than a defined set of relevant views, either because the decoder weights the relevant views more effectively or because it extracts complementary information from views not considered relevant to a given diagnosis. In practice, \echotab{} can therefore be applied directly to all views of a study, without any per-diagnosis view curation.
\section{Dataset Details of Diagnosis Classification Tasks}
\label{app:diagnosis_clf_dataset_details}

We evaluate on a set of ordinal severity-grading tasks derived from the diagnostic labels in the structured measurements, which were held out during measurement selection. The labels in our internal cohort use a four-level scale (normal, mild, moderate, severe), while the labels in the MIMIC dataset use a seven-level scale that additionally distinguishes trace and the intermediate mild-moderate and moderate-severe grades. Studies without a label for a given task are excluded from that task, so per-task totals differ within a split.

\Cref{tab:diagnosis_class_distribution_mimic} and \Cref{tab:diagnosis_class_distribution_internal} give the class distributions for MIMIC and our internal cohort, respectively. For training, we balance the classes as much as possible so that the $k$-NN classifier sees roughly an equal number of samples from each class when making predictions. The validation and test sets retain the original class distributions. Meanwhile, \Cref{tab:diagnosis_measurement_view_mapping_mimic} and \Cref{tab:diagnosis_measurement_view_mapping_internal} show the diagnosis-to-measurement-to-view mappings for MIMIC and our internal dataset, respectively, as defined by a cardiologist and used to identify diagnosis-relevant measurements and views.

\begin{table}[h]
  \centering
  \scriptsize
  \setlength{\tabcolsep}{4pt}
  \caption{Class distribution of the downstream diagnosis classification tasks of the MIMIC dataset
  (\emph{N}ormal, \emph{T}race, \emph{Mi}ld, \emph{Mi-Mo} mild-to-moderate,
  \emph{Mo}derate, \emph{Mo-S} moderate-to-severe, \emph{S}evere).}
  \label{tab:diagnosis_class_distribution_mimic}
  \begin{tabular}{lrrrrrrr}
    \toprule
    \textbf{Task} & \textbf{N} & \textbf{T} & \textbf{Mi} & \textbf{Mi-Mo} & \textbf{Mo} & \textbf{Mo-S} & \textbf{S} \\
    \midrule
    \multicolumn{8}{l}{\emph{Train}} \\
    Aortic stenosis (AS)       &  115 &    - &  112 &   - & 114 &   - & 112 \\
    Aortic regurgitation (AR)   &    - & 9 &  8 & 9 &  9 &  9 &   9 \\
    Mitral regurgitation (MR)    &   73 & 75 & 74 & 75 & 74 & 73 &  74 \\
    Left atrium size (LA) & 417 &    - &  413 &   - & 417 &   - &   - \\
    Right atrium size (RA)  & 152 &    - & 153 &   - & 152 &   - & 149 \\
    Left ventricle size (LV) & 5 &    - &   5 &   - &  5 &   - &   5 \\
    Right ventricle size (RV) & 29 &    - &  29 &   - & 29 &   - &  29 \\
    Left ventricle ejection fraction (LVEF) &  40 &    - &  38 &   - &  38 &   - &  40 \\
    Right ventricle ejection fraction (RVEF) & 87 &    - &  87 &   - & 85 &   - &  86 \\
    
    \midrule
    
    \multicolumn{8}{l}{\emph{Val}} \\
    Aortic stenosis (AS)        &  12 &   - &  22 &   - &  20 &  - & 22 \\
    Aortic regurgitation (AR)   &   - & 136 & 102 &  20 &  12 &  2 &  - \\
    Mitral regurgitation (MR)    &   9 & 300 & 195 &  61 &  56 & 25 &  7 \\
    Left atrium size (LA) & 243 &   - &  72 &   - &  58 &  - &  - \\
    Right atrium size (RA) & 325 &   - & 177 &   - &  38 &  - & 17 \\
    Left ventricle size (LV) & 175 &   - &   9 &   - &   8 &  - &  - \\
    Right ventricle size (RV) & 531 &   - & 100 &   - &  17 &  - &  4 \\
    Left ventricle ejection fraction (LVEF) & 100 &  - &  31 &   - &   2 &  - & 10 \\
    Right ventricle ejection fraction (RVEF) & 577 &   - &  53 &   - &  24 &  - & 15 \\

    \midrule
    
    \multicolumn{8}{l}{\emph{Test}} \\
    Aortic stenosis (AS) &  20 &   - &  33 &   - &   9 &  - & 19 \\
    Aortic regurgitation (AR) &   - & 134 & 114 &  24 &  14 &  1 &  2 \\
    Mitral regurgitation (MR) &   7 & 338 & 176 &  61 &  54 & 30 &  5 \\
    Left atrium size (LA) & 254 &   - &  92 &   - &  50 &  - &  - \\
    Right atrium size (RA) & 346 &   - & 172 &   - &  33 &  - & 23 \\
    Left ventricle size (LV) & 216 &   - &   5 &   - &  14 &  - &  3 \\
    Right ventricle size (RV) & 508 &   - & 113 &   - &  41 &  - &  2 \\
    Left ventricle ejection fraction (LVEF) &  78 &   - &  40 &   - &   5 &  - &  3 \\
    Right ventricle ejection fraction (RVEF) & 555 &   - &  67 &   - &  35 &  - & 16 \\
    
    \bottomrule
  \end{tabular}
\end{table}

\begin{table}[h]
  \centering
  \scriptsize
  \setlength{\tabcolsep}{4pt}
  \caption{Class distribution of the downstream diagnosis classification tasks of our internal dataset
  (\emph{N}ormal, \emph{Mi}ld, \emph{Mo}derate, \emph{S}evere).}
  \label{tab:diagnosis_class_distribution_internal}
  \begin{tabular}{lrrrrcrrrrcrrrr}
    \toprule
    
    & \multicolumn{4}{c}{\textbf{Train}} & & \multicolumn{4}{c}{\textbf{Val}} & & \multicolumn{4}{c}{\textbf{Test}} \\
    \cmidrule(lr){2-5} \cmidrule(lr){7-10} \cmidrule(lr){12-15}
    \textbf{Task} & \textbf{N} & \textbf{Mi} & \textbf{Mo} & \textbf{S} & & \textbf{N} & \textbf{Mi} & \textbf{Mo} & \textbf{S} & &  \textbf{N} & \textbf{Mi} & \textbf{Mo} & \textbf{S} \\
    
    \midrule

    Aortic stenosis (AS)        & 96 & 96 & 96 & 96 & & 724 &  28 & 21 & 10 & & 736 &  21 & 22 & 20 \\
    Mitral regurgitation (MR)    & 37 & 37 & 37 & 37 & & 557 & 176 & 40 & 10 & & 552 & 202 & 35 & 10 \\
    Left atrium size (LA)                 & 113 & 113 & 113 & 113 & & 626 &  47 & 79 & 26 & & 658 &  46 & 58 & 24 \\
    Right atrium size (RA)                & 73 &  73 & 73 &  73 & & 704 &  29 & 27 & 19 & & 738 &  15 & 21 & 17 \\
    Left ventricle size (LV)                 & 22 &  22 &  22 &  22 & & 714 &  12 & 19 &  9 & & 711 &  14 & 14 &  6 \\
    Right ventricle size (RV)                 & 14 & 14 & 14 & 14 & & 721 &  26 &  8 &  - & & 716 &  17 & 12 &  - \\
    Left ventricle ejection fraction (LVEF)              & 109 & 109 & 109 & 109 & & 633 &  62 & 53 & 43 & & 634 &  71 & 55 & 47 \\
    Right ventricle ejection fraction (RVEF)            & 13 &  13 &  13 &  13 & & 751 &  22 &  6 &  3 & & 770 &  16 & 11 &  - \\
    
    \bottomrule
  \end{tabular}
\end{table}

\begin{table}[t]
  \centering
  \caption{Diagnosis $\rightarrow$ Measurement $\rightarrow$ View mapping for the MIMIC dataset, as defined by a cardiologist.}
  \label{tab:diagnosis_measurement_view_mapping_mimic}
  \scriptsize
  \begin{tabularx}{\textwidth}{@{} l >{\raggedright\arraybackslash}X l @{}}
    \toprule
    \textbf{Diagnosis} & \textbf{Measurements} & \textbf{Views} \\
    \midrule
    Aortic regurgitation (AR)
      & \texttt{ar\_pr\_halftime}
      & Apical Doppler \\
    \addlinespace
    Aortic stenosis (AS)
      & \texttt{av\_mean\_grad}, \texttt{av\_pk\_grad}, \texttt{av\_pk\_vel}, \texttt{av\_area\_continuity}, \texttt{av\_vti}, \texttt{lvot\_diam}, \texttt{lvot\_pk\_vel}, \texttt{lvot\_vti}
      & Apical Doppler, PLAX \\
    \addlinespace
    Mitral regurgitation (MR)
      & \texttt{mv\_e\_decel}
      & Apical Doppler \\
    \addlinespace
    Left atrium size (LA)
      & \texttt{la\_dimen}, \texttt{la\_4ch\_length}, \texttt{la\_vol}
      & PLAX, A4C, A2C \\
    \addlinespace
    Right atrium size (RA)
      & \texttt{ra\_length}
      & A4C \\
    \addlinespace
    Left ventricle size (LV)
      & \texttt{lvedd}, \texttt{lvesd}, \texttt{biplane\_lvedv}, \texttt{biplane\_lvesv}
      & PLAX, A2C, A4C \\
    \addlinespace
    Right ventricle size (RV)
      & \texttt{rv\_diam}
      & A4C \\
    \addlinespace
    Left ventricle ejection fraction (LVEF)
      & \texttt{biplane\_lvef}
      & A2C, A4C \\
    \addlinespace
    Right ventricle ejection fraction (RVEF)
      & \texttt{tapse}
      & A4C \\
    \bottomrule
  \end{tabularx}
\end{table}

\begin{table}[t]
  \centering
  \caption{Diagnosis $\rightarrow$ Measurement $\rightarrow$ View mapping for our internal dataset, as defined by a cardiologist.}
  \label{tab:diagnosis_measurement_view_mapping_internal}
  \scriptsize
  \begin{tabularx}{\textwidth}{@{} l >{\raggedright\arraybackslash}X l @{}}
    \toprule
    \textbf{Diagnosis} & \textbf{Measurements} & \textbf{Views} \\
    \midrule
    Aortic stenosis (AS)
      & \texttt{AV\_Pmean}, \texttt{AV\_Pmax}, \texttt{AV\_Vmax}, \texttt{AV\_VTI}, \texttt{AV\_COA\_Vmax}, \texttt{LVOT\_Pmax}, \texttt{LVOT\_Pmean}, \texttt{LVOT\_VTI}, \texttt{LVOT\_SV\_D}, \texttt{LVOT\_Vmax}, \texttt{LVOT\_Vmean}, \texttt{LVOTd\_2D}
      & Apical Doppler, PLAX \\
    \addlinespace
    Mitral regurgitation (MR)
      & \texttt{MV\_DT}
      & Apical Doppler \\
    \addlinespace
    Left atrium size (LA)
      & \texttt{LADs\_AC4}, \texttt{LAESV\_MODBP}, \texttt{LAAs\_MOD4}
      & PLAX, A4C, A2C \\
    \addlinespace
    Right atrium size (RA)
      & \texttt{RAA\_2D}, \texttt{RAD\_l\_2D}, \texttt{RAV\_A4C}
      & A4C \\
    \addlinespace
    Left ventricle size (LV)
      & \texttt{LVDd\_2D}, \texttt{LVDs\_2D}, \texttt{LVEDV\_MODBP}, \texttt{LVESV\_MODBP}
      & PLAX, A2C, A4C \\
    \addlinespace
    Right ventricle size (RV)
      & \texttt{RVDd\_2D}
      & A4C \\
    \addlinespace
    Left ventricle ejection fraction (LVEF)
      & \texttt{LVEF\_MODBP}, \texttt{LVSV\_MODBP}
      & A2C, A4C \\
    \addlinespace
    Right ventricle ejection fraction (RVEF)
      & \texttt{TAPSE}
      & A4C \\
    \bottomrule
  \end{tabularx}
\end{table}

\section{Echos}

\subsection{Echo study distribution}

\begin{figure}[htbp]
    \centering
        \centering
        \includegraphics[width=\linewidth]{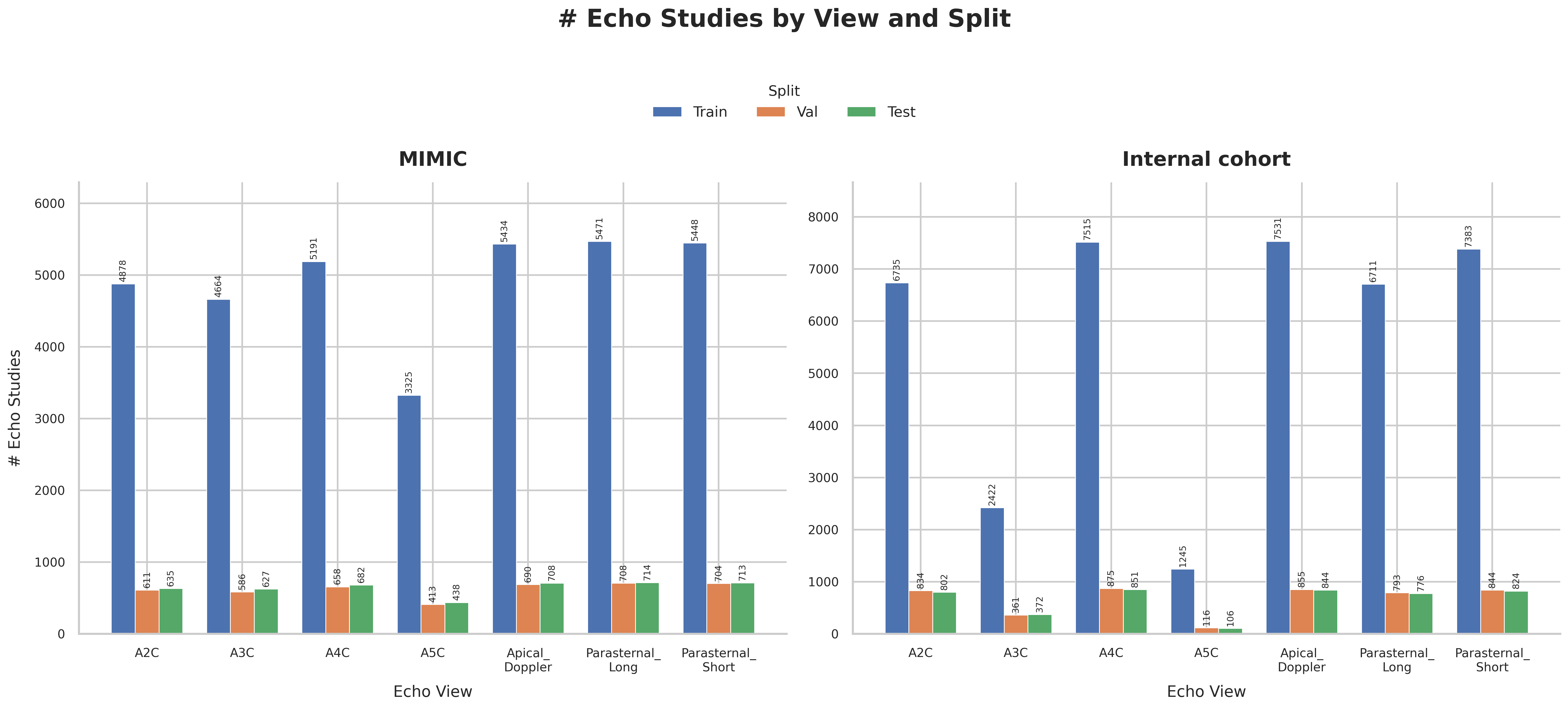}
    \caption{Distribution of echo studies by view and data split (train/val/test) for the MIMIC and internal cohort datasets.}
    \label{fig:echo_study_distribution}
\end{figure}

Figure \ref{fig:echo_study_distribution} shows the distribution of echo studies by echo view and data split (train/val/test) for the MIMIC dataset and our internal cohort.

\subsection{Echo pre-processing}

Each echo video first undergoes pre-processing to isolate the ultrasound region using the \texttt{SequenceOfUltrasoundRegions} field from the DICOM metadata, and to strip out the ECG overlay along with any remaining burned-in text or annotations. The videos are then resized to $224 \times 224$ pixels and stored as AVI files. As in EchoPrime, an additional 10\% zoom crop is applied at data-loading time to cut down on empty background, after which the frames are resized back to $224 \times 224$. Pixel values are normalized based on the mean and standard deviation from the EchoPrime training set. Each video is also sampled down to 16 frames using a stride of 2. Videos that come up short are zero-padded to a uniform length. During training, no augmentation is applied to the echos.

\subsection{Echo Token Sequence}
\label{app:echo_tokens}

Table \ref{tab:echo_view_clf} reports $k$-NN ($k=5$) performance on the view classification task across different configurations of the echo tokens from EchoPrime. Extracting local patch tokens from EchoPrime and subsequently removing those corresponding to background patches yields the best performance, indicating that this configuration produces the most discriminative echo tokens across views. This is a desirable property because we want the measurement queries in \echotab{} to learn to attend to the correct echo views.

\begin{table}[h]
\centering
\footnotesize
\begin{tabular}{cccc}
\toprule
\textbf{Token type} & \textbf{Remove background tokens} & \textbf{AUROC} & \textbf{Accuracy} \\
\midrule
Global (CLS) & - & 0.752 & 0.456 \\
Local & $\times$ & 0.792 & 0.554 \\
Local & $\checkmark$ & \textbf{0.805} & \textbf{0.583} \\
\bottomrule
\end{tabular}
\caption{\textbf{$k$-NN ($k=5$) performance on echo view classification using different configurations of echo tokens from EchoPrime.} We use the configuration in the last row for all models that use EchoPrime: local tokens, excluding those that correspond to the background.}
\label{tab:echo_view_clf}
\end{table}
\section{Measurements}
\label{app:measurements}

\Cref{fig:measurements_mimic} and \Cref{fig:measurements_internal} show the list of measurements used from the MIMIC dataset and our internal dataset, respectively.

\begin{figure}[h]
    \centering
    \begin{tikzpicture}
        \node[
            draw,
            rounded corners,
            text width=0.95\linewidth,
            align=left,
            inner sep=6pt
        ] {
        \small
        \texttt{ar\_pr\_halftime, arch\_diam, ascending\_diam, av\_area\_continuity, av\_mean\_grad, av\_pk\_grad, av\_pk\_vel, av\_vti, biplane\_lvedv, biplane\_lvef, biplane\_lvesv, descending\_diam, e\_e\_prime, fs, inf\_lat\_thickness, ivc\_diam, la\_4ch\_length, la\_dimen, la\_vol, lat\_e\_prime, lateral\_E\_to\_eprime, lvedd, lvesd, lvot\_diam, lvot\_pk\_vel, lvot\_vti, mean\_E\_to\_eprime, mv\_e\_decel, mv\_mean\_grad, mv\_peak\_a, mv\_peak\_e, mv\_peak\_e\_a, mv\_peak\_vel, mva2, pv\_peak\_vel, ra\_length, rest\_lvot, rv\_diam, sept\_e\_prime, septal\_E\_to\_eprime, septal\_thickness, sinus\_diam, tapse, tr\_mmhg, tr\_velocity}
        };
    \end{tikzpicture}
    \caption{List of measurements used from the MIMIC dataset (45 total).}
    \label{fig:measurements_mimic}
\end{figure}

\begin{figure}[h]
    \centering
    \begin{tikzpicture}
        \node[
            draw,
            rounded corners,
            text width=0.95\linewidth,
            align=left,
            inner sep=6pt
        ] {
        \small
        \texttt{AoAsc\_2D, Ao\_MM, AR\_DS, AR\_PHT, AR\_Pmax, AR\_Vmax, AVA\_D, AV\_COA\_Vmax, AV\_Pmax, AV\_Pmean, AV\_VR, AV\_VTI, AV\_Vmax, AV\_Vmean, IVSd\_2D, LAAs\_MOD4, LADs\_AC4, LAESV\_MODBP, LA\_AO\_MM, FS\_2D, IVS\_LVPW\_T, LVAd\_AL4, LVAs\_AC4\_2D, LVDd\_2D, LVDs\_2D, LVEDV\_MODBP, LVEF\_MODBP, LVESV\_MODBP, LVLd\_AL4, LVLs\_AL4, LVM\_MM, LVOTA\_2D, LVOT\_Pmax, LVOT\_Pmean, LVOT\_SV\_D, LVOT\_VTI, LVOT\_Vmax, LVOT\_Vmean, LVOTd\_2D, LVPWd\_2D, LVSV\_MODBP, MVA\_PHT, MV\_A\_Vmax, MV\_DS, MV\_DT, MV\_EA, MV\_E\_Eann, MV\_E\_Eann\_lat, MV\_E\_Vmax, MV\_Eann, MV\_Eann\_lat, MV\_PHT, MV\_Pmax, MV\_Pmean, MV\_VTI, MV\_Vmax, MV\_Vmean, PV\_ACCS, PV\_Acctime, RAA\_2D, RAD\_l\_2D, RAV\_A4C, RVDd\_2D, RV\_LV\_EDD, TAPSE, TR\_Pmax, TR\_Vmax}
        };
    \end{tikzpicture}
    \caption{List of measurements used from our internal dataset (67 total).}
    \label{fig:measurements_internal}
\end{figure}



\section{Hyperparameter Settings}
\label{app:hyperparameters}

\echotab{} is trained for 30 epochs using the AdamW optimizer with a base batch size of 4 and gradient accumulation over 16 iterations. $d$ is set to 768. We use a base learning rate of $1\times10^{-4}$ for the EchoPrime-initialized echo encoder on MIMIC and $5\times10^{-5}$ on our internal dataset, while using $1\times10^{-4}$ for all other trainable parameters, including the tabular decoder, measurement queries, and prediction head. We use a warmup-cosine schedule, linearly increasing the learning rate from 10\% to 100\% of the base rate over two epochs, followed by cosine decay to 10\% by the end of training. The tabular decoder consists of two cross-attention blocks with four attention heads each, while all prediction-head variants use a two-layer MLP. For the partially shared head, we use a low-rank update with rank $r=8$ and scaling factor $\alpha=16$. We apply gradient clipping with a maximum norm of 50.

\section{Hypernetwork Variants}

Inspired by text-to-LoRA \citep{texttolora}, we experimented with three variants of the hypernetwork used in the partially shared prediction head of \echotab{}: \texttt{L}, \texttt{M}, and \texttt{S}. By default, our model uses \texttt{M}.

All variants share the same hypernetwork backbone (\Cref{fig:hypernet_backbone}) and differ only in how they generate the LoRA factors $A_{m,\ell}\in\mathbb{R}^{d\times r}$ and $B_{m,\ell}\in\mathbb{R}^{r\times d}$, with $d=768$ and $r=8$ (\Cref{fig:hypernet_variants}). The largest variant, \texttt{L}, generates both factors $A$ and $B$ in a single forward pass, requiring an output dim. of $2rd=12288$ in \texttt{h1w}. The medium variant, \texttt{M}, instead runs the shared body twice, generating one factor per pass and reducing the output dim. of \texttt{h1w} to $rd=6144$. The two passes are distinguished by a learned $\texttt{ab\_emb}$ embedding, with one embedding vector for each factor, which is added to the conditioning vector $c$. The smallest variant, \texttt{S}, further factorizes generation by producing one rank slice of a single factor at a time. It therefore reduces the output dimension to $d=768$ while requiring $2r=16$ forward passes through the shared body. Each pass is conditioned on both the corresponding $\texttt{ab\_emb}$ and a learned $\texttt{rank\_emb}$, which provides one embedding vector for each rank index. Thus, moving from \texttt{L} to \texttt{S} trades additional computation for  fewer hypernetwork parameters: the number of parameters decreases, while the number of forward passes through the shared body increases. \Cref{tab:ablation_hypernetwork_variants} shows the ablation of these different hypernetwork variants.

\begin{figure}[t]
\hrule\vspace{4pt}
\begin{lstlisting}[style=pytorchrepr]
Hypernet(
  (cond_projection): Sequential(
    (0): Linear(in_features=768, out_features=128, bias=True)
    (1): LayerNorm((128,), eps=1e-05, elementwise_affine=True)
  )
  (layer_idx_emb): Embedding(1, 128)
  (layer_idx_emb_norm): LayerNorm((128,), eps=1e-05, elementwise_affine=True)
  (body): Sequential(
    (0): Linear(in_features=128, out_features=256, bias=True)
    (1): GELU()
    (2): Dropout(p=0.05, inplace=False)
    (3): Linear(in_features=256, out_features=128, bias=True)
    (4): GELU()
  )
  (h1w_head): Linear(in_features=128, out_features=6144, bias=True)
  (h1b_head): Linear(in_features=128, out_features=768, bias=True)
  (h2w_head): Linear(in_features=128, out_features=768, bias=True)
  (h2b_head): Linear(in_features=128, out_features=1, bias=True)
)
\end{lstlisting}
\vspace{-4pt}\hrule
\caption{Hypernetwork that generates the measurement-specific corrections in the partially shared prediction head (pictured here is the \texttt{M} variant). The $d$-d measurement query is projected to a 128-d conditioning vector $c$, passed through the body ($h = \mathrm{body}(c) + c$), and decoded by four linear heads: \texttt{h1w}/\texttt{h1b} produce the rank-8 LoRA factors ($A_{m,\ell}$ and $B_{m,\ell}$) and bias ($\beta_{m,\ell}$) of the hidden layer $\ell$ of shape $d \times d$, and \texttt{h2w}/\texttt{h2b} the weight delta ($a_{m,L}$) and bias ($\beta'_{m,L}$) of the final layer $L$ of shape $d \times 1$. Additionally, the \texttt{h1} heads are conditioned on $c + \mathrm{LN}(\texttt{layer\_idx\_emb}[\ell])$, so that it can generate a distinct hidden layer for each depth $\ell$. The \texttt{h2} heads use $c$ directly, as there is only a single final layer $L$. All generated parameters are added to the base MLP shared across measurements.}
\label{fig:hypernet_backbone}
\end{figure}

\begin{figure}[h]
\hrule\vspace{4pt}
\begin{lstlisting}[style=pytorchrepr]
# L
(h1w_head): Linear(in_features=128, out_features=12288, bias=True)

# M
(ab_emb):      Parameter (2, 128)
(ab_emb_norm): LayerNorm((128,), eps=1e-05, elementwise_affine=True)
(h1w_head):    Linear(in_features=128, out_features=6144, bias=True)

# S
(ab_emb):        Parameter (2, 128)
(ab_emb_norm):   LayerNorm((128,), eps=1e-05, elementwise_affine=True)
(rank_emb):      Parameter (8, 128)
(rank_emb_norm): LayerNorm((128,), eps=1e-05, elementwise_affine=True)
(h1w_head):      Linear(in_features=128, out_features=768, bias=True)
\end{lstlisting}
\vspace{-4pt}\hrule
\vspace{6pt}
\centering
{\footnotesize
\begin{tabular}{lcc}
\toprule
\textbf{Hypernetwork variant} & \textbf{\# passes} & \textbf{\texttt{h1w\_head} output dim.} \\
\midrule
\texttt{L} & $1$      & $2rd = 12288$ \\
\texttt{M} & $2$      & $rd = 6144$ \\
\texttt{S} & $2r = 16$ & $d = 768$ \\
\bottomrule
\end{tabular}}
\caption{Hypernetwork variants used in the partially shared prediction head of \echotab{}. Assume $r=8$ and $d=768$. Every other module of the hypernetwork is identical across variants (\Cref{fig:hypernet_backbone}).}
\label{fig:hypernet_variants}
\end{figure}

\begin{table}[h]
\centering
{\footnotesize
\setlength{\tabcolsep}{4pt}
\begin{tabular}{l cc}
\toprule
\textbf{Hypernetwork variant} & \textbf{Diag. clf. (avg. AUC)} & \textbf{Meas. pred. (avg. $R^2$)} \\
\midrule
\texttt{L} & .735 & .405 \\
\texttt{M} & \textbf{.748} & \textbf{.418} \\
\texttt{S} & .743  & .395 \\
\bottomrule
\end{tabular}}
\caption{\textbf{Ablation of the hypernetwork variants used in the partially shared prediction head of \echotab{} on the MIMIC dataset.} By default, our model uses the \texttt{M} variant.}
\label{tab:ablation_hypernetwork_variants}
\end{table}
\section{Baselines}
\label{app:baselines}

This section describes the baselines used in our experiments in more detail: MLP (\Cref{app:mlp_baseline}), contrastive (\Cref{app:contrastive_baseline}), and SupCon (\Cref{app:supcon_baseline}). All baselines use the same EchoPrime-initialized echo encoder and view set $\mathcal{C}$ as \echotab{}.

\subsection{MLP}
\label{app:mlp_baseline}

The MLP baseline adds a separate prediction head for each measurement on top of the echo encoder, analogous to the independent-head variant of \echotab{}. The study-level echo embedding is constructed by averaging over the echo view embeddings. It is then passed to each measurement-specific head, with all heads using the same $L$-layer MLP architecture as \echotab{}'s prediction head.

The MLP baseline is trained for 30 epochs with the AdamW optimizer, using a base batch size of 4 and gradient accumulation over 16 iterations. We use a base learning rate of $1\times10^{-4}$ for both the MIMIC and internal datasets. We use a warmup-cosine schedule, linearly increasing the learning rate from 10\% to 100\% of the base rate over two epochs, followed by cosine decay to 10\% by the end of training. For the echo encoder, we apply a layer-wise learning-rate decay of 0.75 on MIMIC and no decay on the internal dataset.

\subsection{Contrastive}
\label{app:contrastive_baseline}

We align paired echo and measurement representations using a CLIP-style contrastive objective \citep{clip}, following the image--tabular formulation of MMCL \citep{mmcl}. For a batch of $N$ samples, let $\mathbf{e}_i$ denote the echo representation and $\mathbf{t}_i$ the corresponding measurement representation. Both are projected to 128-dim by separate linear projection layers, yielding $\mathbf{z}^{\mathrm{e}}_i$ and $\mathbf{z}^{\mathrm{t}}_i$, respectively. The echo-to-measurement contrastive loss is

\begin{equation}
\mathcal{L}_{\mathrm{E}\rightarrow\mathrm{M}} = -\frac{1}{N}
\sum_{i=1}^{N}
\log
\frac{
\exp\left(
\operatorname{cos}(\mathbf{z}^{\mathrm{e}}_i,\mathbf{z}^{\mathrm{t}}_i)/\tau
\right)
}{
\sum_{j=1}^{N}
\exp\left(
\operatorname{cos}(\mathbf{z}^{\mathrm{e}}_i,\mathbf{z}^{\mathrm{t}}_j)/\tau
\right)
},
\end{equation}

where $\tau$ is the temperature. We set $\tau = 0.07$. We additionally compute the symmetric measurement-to-echo loss $\mathcal{L}_{\mathrm{M}\rightarrow\mathrm{E}}$ by exchanging the two modalities, and optimize a weighted combination of the two directional losses.

We consider two variants that differ in the sample unit. In the \emph{study-level} variant, $\mathbf{e}_i$ is obtained by attention-pooling the available view representations of a study. In the \emph{view-level} variant, $\mathbf{e}_i$ is a view representation and is paired with the measurement representation of its corresponding study. The view-level variant therefore provides substantially more samples, enabling larger batches and more negative pairs.

The measurements for each study are encoded with the TARTE tabular encoder \citep{tarte}. Because TARTE requires column names corresponding to the input values, we use the measurement descriptions provided by MIMIC and our internal dataset as column names. We use the TARTE \texttt{CLS} token as $\mathbf{t}_i$. To maximize batch size, these measurement representations are pre-computed using a pre-trained and frozen TARTE encoder, while the echo encoder is fully fine-tuned.

We train for 30 epochs using the AdamW optimizer with a batch size of 192 for the view-level variant and 32 for the study-level variant. The base learning rate is $5\times10^{-5}$ for the study-level variant on both datasets and $1\times10^{-4}$ for the view-level variant on both datasets. We use a warmup-cosine schedule, linearly increasing the learning rate from 10\% to 100\% of the base rate over two epochs, followed by cosine decay to 10\% by the end of training.


\subsection{Supervised Contrastive (SupCon)}
\label{app:supcon_baseline}

The SupCon baseline uses tabular measurement similarity to define positive and negative echo pairs, following TGV \citep{tgv} and inspired by the original SupCon \citep{khosla2020supervised}. Each measurement is $z$-score normalized using training-set statistics, and missing values are mean-imputed, corresponding to zero after normalization.

For a batch of $N$ samples, let $\mathbf{e}_i$ denote the echo representation after projection by a linear layer and $\mathbf{x}_i\in\mathbb{R}^{F}$ the corresponding vector of $F$ normalized measurements. We $\ell_2$-normalize the representations:

\begin{equation}
\hat{\mathbf{e}}_i
=
\frac{\mathbf{e}_i}{|\mathbf{e}_i|_2},
\qquad
\hat{\mathbf{x}}_i
=
\frac{\mathbf{x}_i}{|\mathbf{x}_i|_2}.
\end{equation}

The tabular similarity between samples $i$ and $j$ is then

\begin{equation}
S_{ij}
=
\hat{\mathbf{x}}_i^{\top}\hat{\mathbf{x}}_j.
\end{equation}

Following TGV, positives are defined relative to each anchor rather than using a fixed similarity threshold. For anchor $i$, let

\begin{equation}
M_i = \max_{j\neq i} S_{ij},
\end{equation}

and define the positive set as

\begin{equation}
    \mathcal{P}(i) = \{\, j \neq i \mid S_{ij} \geq M_i - h \,\}.
\end{equation}

which guarantees at least one positive for every anchor. All remaining samples in the batch serve as negatives. The training objective is

\begin{equation}
\mathcal{L}
=
-\frac{1}{N}
\sum_{i=1}^{N}
\log
\frac{
\sum_{j\in\mathcal{P}(i)}
\exp\left(
\hat{\mathbf{e}}_i^{\top}\hat{\mathbf{e}}_j/\tau
\right)
}{
\sum_{k=1}^{N}
\exp\left(
\hat{\mathbf{e}}_i^{\top}\hat{\mathbf{e}}_k/\tau
\right)
},
\end{equation}

where $\tau$ is the temperature. We set $\tau=0.07$ and $h=0.05$.

As in the contrastive baseline, we consider both \emph{study-level} and \emph{view-level} variants, using the same definitions of the sample unit. We train for 30 epochs using the AdamW optimizer with a batch size of 192 for the view-level variant and 32 for the study-level variant. The base learning rate is $1\times10^{-4}$ for the study-level variant and $1\times10^{-4}$ for the view-level variant on MIMIC, while it is $5\times10^{-5}$ for both variants on our internal dataset. We use a warmup-cosine schedule, linearly increasing the learning rate from 10\% to 100\% of the base rate over two epochs, followed by cosine decay to 10\% by the end of training.

\section{Detailed Results on Measurement Prediction}

\begin{figure}[h]
    \centering
        \centering
        \includegraphics[width=0.7\linewidth]{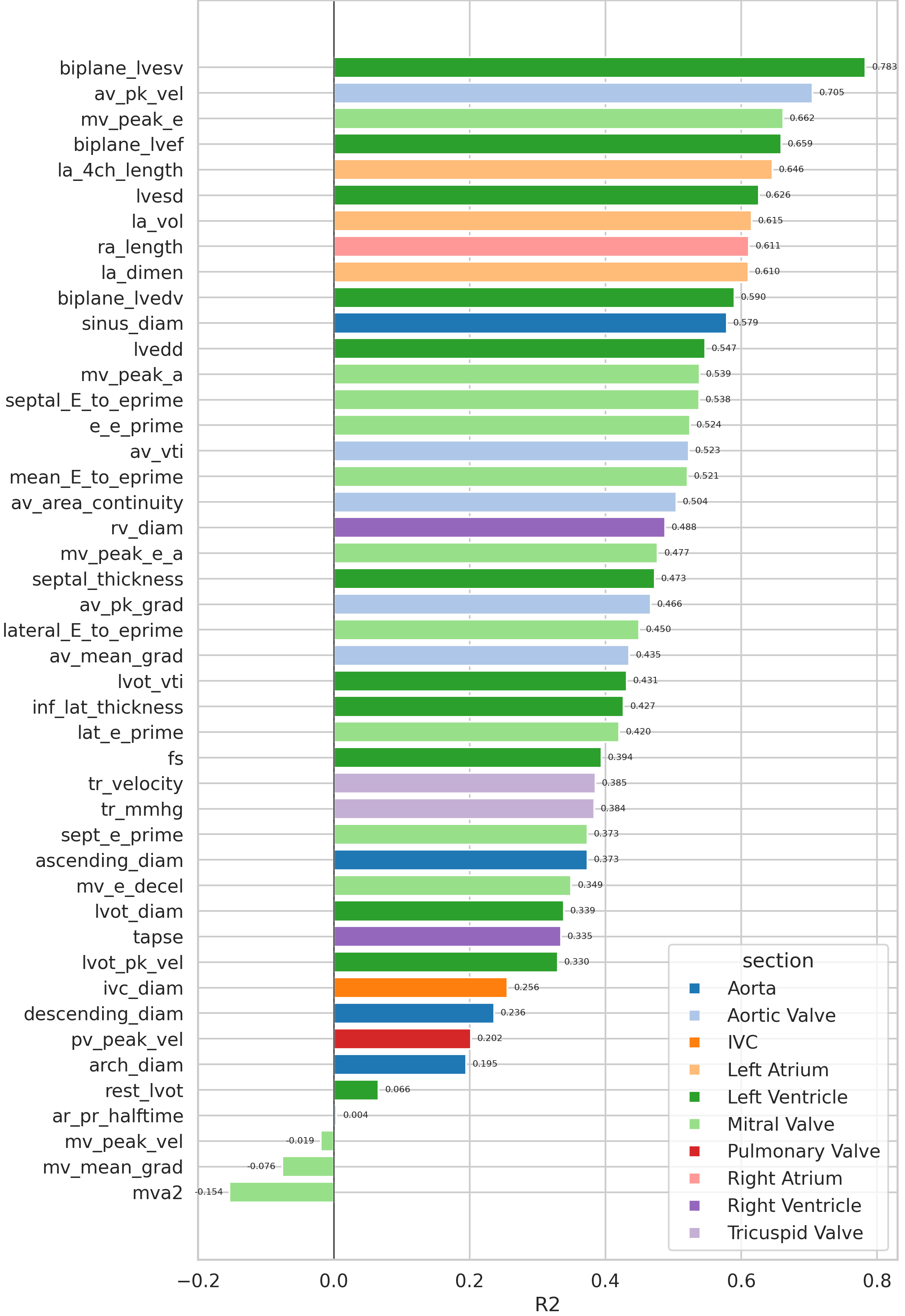}
    \caption{Per-measurement $R^2$ by \echotab{} on the MIMIC dataset.}
    \label{fig:meas_pred_r2_per_meas}
\end{figure}

\begin{figure}[h]
    \centering
        \centering
        \includegraphics[width=0.7\linewidth]{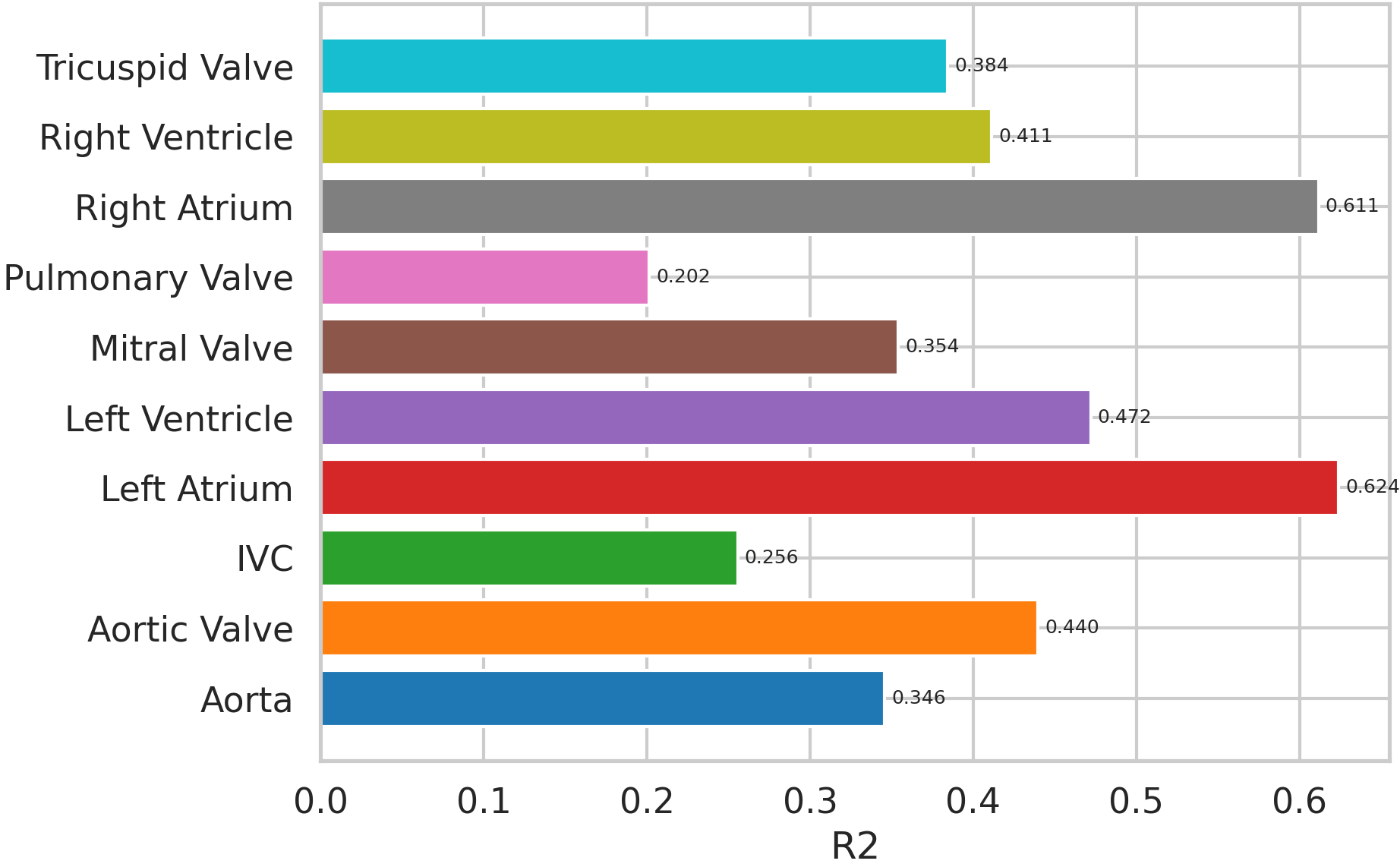}
    \caption{Per-section $R^2$ by \echotab{} on the MIMIC dataset.}
    \label{fig:meas_pred_r2_per_section}
\end{figure}

\Cref{fig:meas_pred_r2_per_meas} and \Cref{fig:meas_pred_r2_per_section} show the per-measurement and per-section $R^2$ by \echotab{} with a partially shared head on the MIMIC dataset.

\end{document}